\documentclass[letterpaper,twocolumn,10pt]{article}

\usepackage{usenix-2020-09}
\usepackage{prp-macros}
\usepackage{amsfonts}
\usepackage{hyperref}
\usepackage{sidecap}
\usepackage{float}
\usepackage{balance} 
\usepackage{amsmath}
\usepackage{algorithm}
\usepackage[noend]{algpseudocode}
\usepackage{blindtext}
\usepackage{amsthm}
\usepackage{amsmath}
\usepackage{microtype}
\usepackage{booktabs}
\usepackage{listings}
\usepackage[english]{babel}
\usepackage{multirow}
\usepackage{wrapfig}
\usepackage{url}
\usepackage[utf8]{inputenc}
\usepackage[english]{babel}
\usepackage{tikz}
\usetikzlibrary{automata, positioning, arrows}
\usepackage{verbatim}
\usepackage{mdframed}
\usepackage{diagbox}
\usepackage[normalem]{ulem}
\usepackage{cleveref}
\usepackage{bm}
\usepackage[symbol]{footmisc}

\usepackage[small,compact]{titlesec}

\usepackage{filecontents}

\newcommand{\sys}{MSRL\xspace} 
\newcommand{\fdg}{FDG\xspace}

\newcommand{\tsc}[1]{\textsuperscript{#1}} 

\renewcommand{\thefootnote}{\fnsymbol{footnote}}

\begin{document}

\date{}

\title{\sys: Distributed Reinforcement Learning with Dataflow Fragments}

\author{
{\rm Huanzhou Zhu\tsc{*}}\\
Imperial College London
\and
{\rm Bo Zhao\tsc{*}}\\
Imperial College London
\and
{\rm Gang Chen}\\
Huawei Technologies Co., Ltd.
\and
{\rm Weifeng Chen}\\
Huawei Technologies Co., Ltd.
\and
{\rm Yijie Chen}\\
Huawei Technologies Co., Ltd.
\and
{\rm Liang Shi}\\
Huawei Technologies Co., Ltd.
\and
{\rm Yaodong Yang}\\
Huawei R\&D United Kingdom
\and
{\rm Peter Pietzuch}\\
Imperial College London
\and
{\rm Lei Chen}\\
Hong Kong University of Science and Technology
} 

\maketitle


\begin{abstract}  
  Reinforcement learning~(RL) trains many agents, which is resource-intensive and must scale to large GPU clusters. Different RL training algorithms offer different opportunities for distributing and parallelising the computation. Yet, current distributed RL systems tie the definition of RL algorithms to their distributed execution: they hard-code particular distribution strategies and only accelerate specific parts of the computation (\eg policy network updates) on GPU workers. Fundamentally, current systems lack abstractions that decouple RL algorithms from their execution. 

  We describe \emph{MindSpore Reinforcement Learning} ({\sys}), a distributed RL training system that supports \emph{distribution policies} that govern how RL training computation is parallelised and distributed on cluster resources, without requiring changes to the algorithm implementation. \sys introduces the new abstraction of a \emph{fragmented dataflow graph}, which maps Python functions from an RL algorithm's training loop to parallel computational \emph{fragments}. Fragments are executed on different devices by translating them to low-level dataflow representations, \eg computational graphs as supported by deep learning engines, CUDA implementations or multi-threaded CPU processes. We show that \sys subsumes the distribution strategies of existing systems, while scaling RL training to 64~GPUs.
  
\end{abstract}


\section{Introduction}
\label{sec:introduction}

\footnotetext[1]{Equal contribution}

Reinforcement learning~(RL) solves decision-making problems in which agents continuously learn policies, typically represented as deep neural networks~(DNNs), on how to act in an unknown environment~\cite{sutton2018reinforcement}.
RL has achieved remarkable outcomes in complex real-world settings: in game play, AlphaGo~\cite{DBLP:journals/nature/SilverHMGSDSAPL16} has defeated a world champion in the board game Go; in biology, AlphaFold~\cite{AlphaFold2021} has predicted three-dimensional structures for protein folding; in robotics, RL-based approaches have allowed robots to perform tasks such as dexterous manipulation without human intervention~\cite{DBLP:conf/icra/0004YZKRXDL21}.

The advances in RL come with increasing computational demands for training: AlphaStar trained 12~agents using 384~TPUs and 1,800~CPUs for more than 44~days to achieve grandmaster level in StarCraft II game play~\cite{DBLP:journals/nature/VinyalsBCMDCCPE19}; OpenAI Five trained to play Dota 2 games for 10~months with 1,536~GPUs and 172,800~CPUs, defeating 99.4\% of human players~\cite{DBLP:journals/corr/abs-1912-06680}. Distributed RL systems must therefore scale in terms of agent numbers, the amount of training data generated by environments, and the complexity of the trained policies~\cite{DBLP:journals/corr/abs-1912-06680, DBLP:journals/nature/VinyalsBCMDCCPE19, AlphaFold2021}.

Existing designs for distributed RL systems, \eg SEED RL~\cite{DBLP:conf/iclr/EspeholtMSWM20}, ACME~\cite{DBLP:journals/corr/abs-2006-00979}, Ray~\cite{DBLP:conf/osdi/MoritzNWTLLEYPJ18}, RLlib~\cite{DBLP:conf/icml/LiangLNMFGGJS18} and Podracer~\cite{DBLP:journals/corr/abs-2104-06272} hardcode a single strategy to parallelise and distribute an RL training algorithm based on its algorithmic structure. For example, if an RL algorithm is defined as a set of Python functions for agents, learners and environments, a system may directly invoke the implementation of an agent's \code{act()} function to generate new actions for the environment.

While integrating an RL algorithm's definition with its execution strategy simplifies the design and implementation, it means that systems fail to parallelise and distribute RL algorithms in the most effective way:

\tinyskip

\noindent

(1)~When \emph{parallelising} the training computation, most RL systems only accelerate the DNN computation on GPUs or TPUs~\cite{DBLP:conf/iclr/EspeholtMSWM20, DBLP:journals/corr/abs-2006-00979, DBLP:conf/icml/LiangLNMFGGJS18} using current deep learning~(DL) engines, such as PyTorch~\cite{DBLP:conf/nips/PaszkeGMLBCKLGA19}, TensorFlow~\cite{TFAgents} or MindSpore~\cite{mindspore}. Other parts of the RL algorithm, such as action generation, environment execution and trajectory sampling, are left to be executed as sequential Python functions on worker nodes, potentilly becoming performance bottlenecks.

Recently, researchers have investigated this unrealised acceleration potential: Podracer~\cite{DBLP:journals/corr/abs-2104-06272} uses the JAX~\cite{jax} compilation framework to vectorise the Python implementation of RL algorithms and parallelise execution on GPUs and TPUs; WarpDrive~\cite{DBLP:journals/corr/abs-2108-13976} can execute the entire RL training loop on a GPU when implemented in CUDA; and RLlib Flow~\cite{DBLP:journals/corr/abs-2011-12719} uses parallel dataflow operators~\cite{DBLP:journals/cacm/ZahariaXWDADMRV16} to express RL training. While these approaches accelerate a larger portion of RL algorithms, they force users to define algorithms through custom APIs \eg with a fixed set of supported computational operators.

\tinyskip

\noindent
(2)~For the \emph{distribution} of the training computation, current RL systems are designed with particular strategies in mind, \ie they allocate algorithmic components (\eg actors and learners) to workers in a fixed way. For example, SEED RL~\cite{DBLP:conf/iclr/EspeholtMSWM20} assumes that learners perform both policy inference and training on TPU cores, while actors execute on CPUs; ACME~\cite{DBLP:journals/corr/abs-2006-00979} only distributes actors and maintains a single learner; and TLeague~\cite{DBLP:journals/corr/abs-2011-12895} distributes learners but co-locates environments with actors on CPU workers. Different RL algorithms deployed in different resources will exhibit different computational bottlenecks, which means that a single strategy for distribution cannot be optimal.

\tinyskip

\noindent
We observe that the above challenges originate from the lack of an \emph{abstraction} that separates the \emph{specification} of an RL algorithm from its \emph{execution}. Inspired by how compilation-based DL engines use intermediate representations~(IRs) to express computation~\cite{xla, DBLP:conf/osdi/ChenMJZYSCWHCGK18}, we want to design a new distributed RL system that (i)~supports the execution of arbitrary parts of the RL computation on parallel devices, such as GPUs and CPUs~(\emph{parallelism}); and (ii)~offers flexibility how to deploy parts of an RL algorithm across devices~(\emph{distribution}).

We describe \emph{MindSpore Reinforcement Learning} ({\sys}), a distributed RL system that decouples the specification of RL algorithms from their execution. This enables \sys to change how it parallelises and distributes different RL algorithms to reduce iteration time and increase the scalability of training. To achieve this, \sys{}'s design makes the following contributions:

\mypar{(1)~Execution-independent algorithm specification~(\S\ref{sec:api})} In \sys, users specify an RL algorithm by implementing its \emph{algorithmic components} (\eg agents, actors, learners) as Python functions in a traditional way. This implementation makes no assumptions about how the algorithm will be executed: all runtime interactions between components are managed by calls to \sys APIs. A separate \emph{deployment configuration} defines the devices available for execution. 

\mypar{(2)~Fragmented dataflow graph abstraction~(\S\ref{sec:dataflow_model})} From the RL algorithm implementation, \sys constructs a \emph{fragmented dataflow graph~(\fdg)}. An \fdg is a dataflow representation of the computation that allows \sys to map the algorithm to heterogeneous devices (CPUs and GPUs) at deployment time. The graph consists of \emph{fragments}, which are parts of the RL algorithm that can be parallelised and distributed across devices.

\sys uses static analysis on the algorithm implementation to group its functions into fragments, instances of which can be deployed on devices. The boundaries between fragments are decided with the help of user-provided \emph{partition annotations} in the algorithm implementation. Annotations specify synchronisation points that require communication between fragments that are replicated across devices for increased data-parallelism.

\mypar{(3)~Fragment execution using DL engines~(\S\ref{sec:design})} For execution, \sys deploys hardware-specific implementations of fragments, permitting instances to run on CPUs and/or GPUs. \sys supports different fragment implementations: CPU implementations use regular Python code, and GPU implementations are generated as compiled computation graphs of DL engines (\eg MindSpore or TensorFlow) if a fragment is implemented using operators, or they are implemented directly in CUDA. \sys fuses multiple data-parallel fragments for more efficient execution.

\mypar{(4)~Distribution policies~(\S\ref{sec:distribution})} Since the algorithm implementation is separated from execution through the \fdg, \sys can apply different \emph{distribution policies} to govern how fragments are mapped to devices. \sys supports a flexible set of distribution policies, which subsume the hard-coded distribution strategies of current RL systems: \eg a distribution policy can distribute multiple actors to scale the interaction with the environment (like Acme~\cite{DBLP:journals/corr/abs-2006-00979}); distribute actors but move policy inference to learners (like SEED RL~\cite{DBLP:conf/iclr/EspeholtMSWM20}); distribute both actors and learners (like Sebulba~\cite{DBLP:journals/corr/abs-2104-06272}); and represent the full RL training loop on a GPU (like WarpDrive~\cite{DBLP:journals/corr/abs-2108-13976} and Anakin~\cite{DBLP:journals/corr/abs-2104-06272}). As the algorithm configuration, its hyper-parameters or hardware resources change, \sys can switch between distribution policy to maintain high training efficiency without requiring changes to the RL algorithm implementation.  

\tinyskip

\noindent
We evaluate \sys experimentally and show that, by switching between distribution policies, \sys improves the training time of the PPO RL algorithm by up 2.4$\times$ as hyper-parameters, network properties or hardware resources change. Its FDG abstraction offers more flexible execution without compromising training performance: \sys scales to 64~GPUs and outperforms the Ray distributed RL system~\cite{DBLP:conf/osdi/MoritzNWTLLEYPJ18} by up to 3$\times$.



\section{Distributed Reinforcement Learning}
\label{sec:background}

Next we give background on reinforcement learning~(\S\ref{sec:background:rl}), discuss the requirements for distributed RL training~(\S\ref{sec:background:dist_rl}), and survey the design space of existing RL systems~(\S\ref{sec:background:design_space}).

\subsection{Reinforcement learning}
\label{sec:background:rl}

\begin{figure}[tb]
  \centering
  \includegraphics[width=\columnwidth]{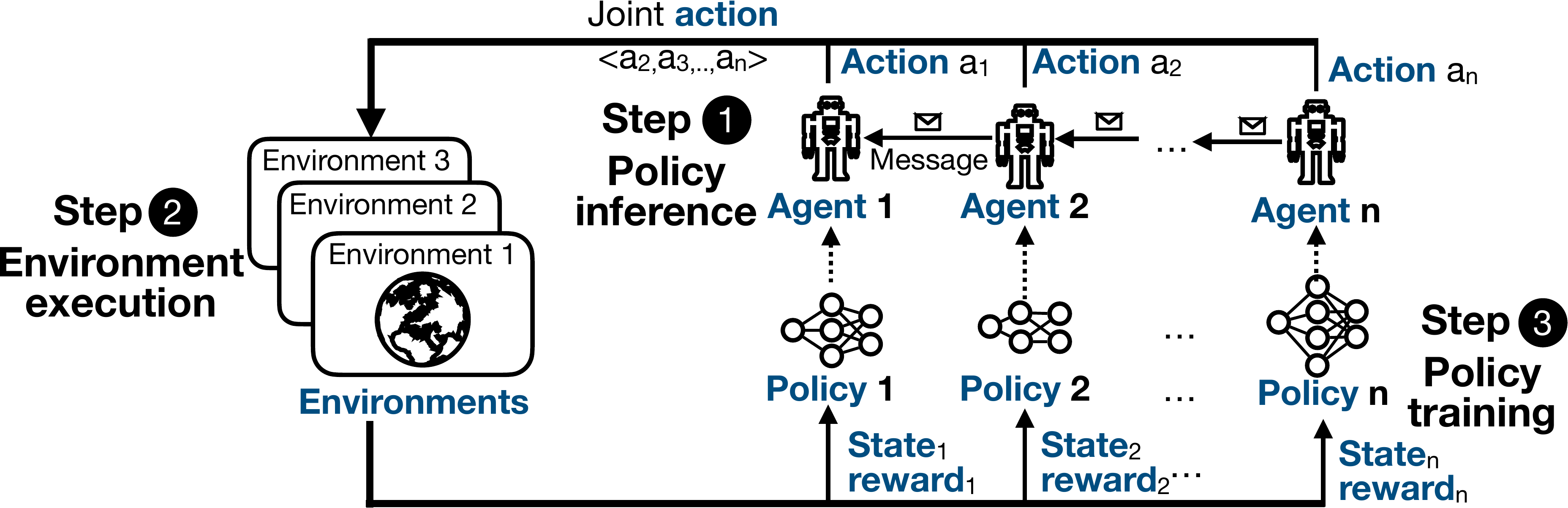}
  \caption{RL training loop with multiple agents}\label{fig:MARLTrain} 
\end{figure}

\emph{Reinforcement learning~(RL)} solves a sequential decision-making problem in which an \emph{agent} operates in an unknown \emph{environment}. The agent's goal is to learn a \emph{policy} that maximises the cumulative \emph{reward} based on the feedback from the environment (see~\F\ref{fig:MARLTrain}): \myc{1} \emph{policy inference}: an agent performs an inference computation on a policy, which maps the environment's state to an agent's action; \myc{2} \emph{environment execution}: the environment executes the actions, generating \emph{trajectories}, which are sequences of $\left\langle \mathit{state}, \mathit{reward} \right\rangle$ pairs produced by the policy; \myc{3} \emph{policy training}: finally, the agent improves the policy by training it based on the received reward.

RL algorithms that train a policy fall into three categories: (1)~\emph{value-based} methods (\eg DQN~\cite{DBLP:journals/nature/MnihKSRVBGRFOPB15}) use a deep neural network~(DNN) to approximate the value function, \ie a mapping of the expected return to perform a given action in a state. Agents learn the values of actions and select actions based on these estimated values; (2)~\emph{policy-based} methods (\eg Reinforce~\cite{DBLP:journals/ml/Williams92}) directly learn a parametrised policy, approximated by a DNN, for selecting actions without consulting a value function. Agents use batched trajectories to train the policy by updating the parameters in the direction of the gradient that maximises the reward; and (3)~\emph{actor–critic} methods (\eg PPO~\cite{DBLP:journals/corr/SchulmanWDRK17}, DDPG~\cite{DBLP:journals/corr/LillicrapHPHETS15}, A2C~\cite{DBLP:conf/icml/MnihBMGLHSK16}) combine the two by learning a policy to select actions (\emph{actor}) and a value function to evaluate selected actions (\emph{critic}).

\emph{Multi-agent reinforcement learning~(MARL)} has multiple agents, each optimising its own cumulative reward when interacting with the environment or other agents (see~\F\ref{fig:MARLTrain}). A3C~\cite{DBLP:conf/icml/MnihBMGLHSK16} executes agents asynchronously on separate environment instances; MAPPO~\cite{mappo2021} extends PPO to a multi-agent setting in which agents share a global parametrised policy. During training, each agent maintains a central critic that takes joint information (\eg state/action pairs) as input, while learning a policy that only depends on local state.

\subsection{Requirements for distributed RL systems}
\label{sec:background:dist_rl}

The benefits of RL come at the expense of its computational cost~\cite{DBLP:conf/icml/Obando-CeronC21}. Complex settings and environments require the exploration of large spaces of actions, states and DNN parameters, and these spaces grow exponentially with the number of agents~\cite{DBLP:conf/ijcai/NairTYPM03}.
Therefore, RL systems must be scalable by exploiting both the parallelism of GPU acceleration and a large number of distributed worker nodes.

There exists a range of proposals how RL systems can parallelise and distribute RL training: for single-agent RL, environment execution~(\myc{2} in \F\ref{fig:MARLTrain}), policy inference and training~(\myc{1}+\myc{3}) can be distributed across workers~\cite{DBLP:conf/osdi/MoritzNWTLLEYPJ18, DBLP:journals/corr/abs-2104-06272, DBLP:journals/corr/abs-2006-00979}, potentially using GPUs~\cite{DBLP:conf/iclr/EspeholtMSWM20, DBLP:journals/corr/abs-2104-06272}; for MARL, agents can be distributed~\cite{DBLP:conf/osdi/MoritzNWTLLEYPJ18, DBLP:conf/icml/LiangLNMFGGJS18, DBLP:journals/corr/abs-2011-12719, DBLP:conf/mlsys/SchaarschmidtMF19} and exchange training state using communication libraries~\cite{nccl, DBLP:conf/osdi/MaiLWFBP20}. In general, environment instances can execute in parallel~\cite{DBLP:conf/icml/MnihBMGLHSK16, DBLP:journals/corr/abs-2104-06272} or be distributed~\cite{DBLP:conf/nips/0001JVBRCL20} to speed up execution.

Not a single of the above strategies for parallelising and distributing RL training is optimal, both in terms of achieving the lowest iteration time and having the best scalability, for all possible RL algorithms. For different algorithms, training workloads and hardware resources, the training bottlenecks shift: \eg our measurements show that, for PPO, environment execution (\myc{2}) takes up to $98\%$ of execution time; for MuZero~\cite{muzero2020}, a large MARL algorithm with many agents, environment execution is no longer the bottleneck, and $97\%$ of time is spent on policy inference and training~(\myc{1}+\myc{3}).

Instead of hardcoding a particular approach for parallelising and distributing RL computation, a distributed RL system design should provide the flexibility to change its execution approach based on the workload. This leads us to the following requirements that our design aims to satisfy:

\mypar{(1)~Execution abstraction} The RL system should have a flexible execution abstraction that enables it to parallelise and distribute computation unencumbered by how the algorithm is defined. While such execution abstractions are commonplace in compilation-based DL engines~\cite{jax, DBLP:conf/osdi/ChenMJZYSCWHCGK18, DBLP:journals/corr/abs-1805-00907, DBLP:conf/cc/BrauckmannGEC20}, they do not exist in current RL systems~(see~\S\ref{sec:background:design_space}).

\mypar{(2)~Distribution strategies} The RL system should support different strategies for distributing RL computation across worker nodes. Users should be permitted to specify multiple distribution strategies for a single RL algorithm, switching between them based on the training workload, without having to change the algorithm implementation. Distribution strategies should be applicable across classes of RL algorithms.

\mypar{(3)~Acceleration support} The RL system should exploit the parallelism of devices, such as multi-core CPUs, GPUs or other AI accelerators. It should support the full spectrum from fine-grained vectorised execution to course-grained task parallel tasks on CPUs cores. Acceleration should not be restricted to policy training and inference only~(\myc{1}+\myc{3}) but cover the full training loop, including environment execution~(\myc{2})~\cite{DBLP:journals/corr/abs-2108-13976}.

\mypar{(4)~Algorithm abstraction} Despite decoupling execution from algorithm specification, users expect familiar APIs for defining algorithms around algorithmic components~\cite{DBLP:conf/nips/KondaT99, DBLP:conf/icml/EspeholtSMSMWDF18, gronauer2021multi}, such as agents, actors, learners, policies, environments etc. An RL system should therefore provide standard APIs, \eg defining the main training loop in terms of policy inference~(\myc{1}), environment execution~(\myc{2}) and policy training~(\myc{3}).

\subsection{Design space of existing RL systems}
\label{sec:background:design_space}

\begin{table*}[t]
	\resizebox{\textwidth}{!}{
		\begin{tabular}{llllcl}
                  \toprule
                  
                  \multicolumn{1}{l}{\textbf{Type}} & \multicolumn{1}{l}{\scell[c]{\textbf{System}}} & \multicolumn{1}{l}{\scell[c]{\textbf{Execution}}} & \multicolumn{1}{l}{\scell[c]{\textbf{Distribution}}} & \multicolumn{1}{l}{\scell[c]{\textbf{Acceleration}}} & \multicolumn{1}{l}{\scell[c]{\textbf{Algorithm}}}\\
			
			\midrule
			\multirow{3}{*}{\scell[c]{\textbf{Function-} \\ \textbf{based} }} & SEED RL~\cite{DBLP:conf/iclr/EspeholtMSWM20}  & \multicolumn{1}{l}{\scell[c]{Python functions}} & \multirow{2}{*}{\scell{environment only}} &  \multicolumn{1}{l}{ \multirow{3}{*}{\scell[c]{DNNs}}} & \multirow{2}{*}{\scell[c]{actor/learner/env}}\\  \cline{2-3}
			& Acme~\cite{DBLP:journals/corr/abs-2006-00979} & \multirow{2}{*}{\scell[c]{Python components}} && &    \\ \cline{2-2} \cline{4-4} \cline{6-6}
			
			& RLGraph~\cite{DBLP:conf/mlsys/SchaarschmidtMF19}   & & \multicolumn{1}{l}{\scell[c]{delegated to backend}} &   & \multicolumn{1}{l}{\scell{agent}}   \\ 
			
			\midrule
			
			\multirow{3}{*}{\scell[c]{\textbf{Actor-} \\  \textbf{based}}} & Ray~\cite{DBLP:conf/osdi/MoritzNWTLLEYPJ18}   & \multicolumn{1}{l}{\multirow{3}{*}{\scell[c]{task (stateless)\\actor (stateful)}}} & \multicolumn{1}{l}{\multirow{3}{*}{\scell[c]{local scheduler\\ global scheduler \\ RPC}}} & \multicolumn{1}{l}{ \multirow{3}{*}{\scell[c]{DNNs}}} & \multirow{2}{*}{\scell[c]{Python functions\\with Ray API~\cite{DBLP:conf/osdi/MoritzNWTLLEYPJ18}}}\\ \cline{2-2}
			
			& RLlib~\cite{DBLP:conf/icml/LiangLNMFGGJS18} &&&& \\ \cline{2-2}\cline{6-6}
			
			& MALib~\cite{DBLP:journals/corr/abs-2106-07551} & 
			&
			& & \multicolumn{1}{l}{\scell[c]{agent/actor/learner/env}}\\  
			
			\midrule
			
                  \multirow{5}{*}{\scell[c]{\textbf{Dataflow-}\\ \textbf{based}}} & Podracer~\cite{DBLP:journals/corr/abs-2104-06272}  & \multicolumn{1}{l}{\scell[c]{JIT-compiled\\ by JAX~\cite{jax} }} & \multicolumn{1}{l}{\scell[c]{two hard-coded\\ distribution schemes}} &  \multicolumn{1}{l}{\scell[c]{funcs/DNNs/envs}} & \scell[c]{JAX~\cite{jax} API}\\ \cline{2-6}
                  
			& RLlib Flow~\cite{DBLP:journals/corr/abs-2011-12719} & \scell[c]{predefined\\ dataflow operators} & \multicolumn{1}{l}{\scell[c]{sharded dataflow operators\\ with Ray tasks~\cite{DBLP:conf/osdi/MoritzNWTLLEYPJ18}}} & \multicolumn{1}{l}{\scell[c]{DNNs}} & \scell[c]{Operator API}\\  \cline{2-6}
			
			& WarpDrive~\cite{DBLP:journals/corr/abs-2108-13976} & \multicolumn{1}{l}{\scell[c]{GPU thread blocks}} & \multicolumn{1}{c}{\scell[c]{\tickNo}} & \multicolumn{1}{l}{\scell[c]{CUDA kernels}} & \multicolumn{1}{l}{\scell[c]{CUDA}}\\

			\midrule\midrule
			
			\multicolumn{1}{l}{\scell[c]{\textbf{Fragmented}\\ \textbf{dataflow graph}}} & \textbf{\sys}  & \multicolumn{1}{l}{\scell[c]{heterogeneous \\dataflow fragments}}  & \multicolumn{1}{l}{\scell[c]{dataflow partitioning}} & \multicolumn{1}{l}{\scell[c]{funcs/operators/\\DNNs/envs}} & 
			\multicolumn{1}{l}{\scell[c]{agent/actor/learner/env}}\\ 
			\bottomrule
	\end{tabular}}
	\caption{Design space of distributed RL systems}
	\label{tab:design_space}
\end{table*}


\begin{figure}[tb]
  \centering
  \begin{subfigure}[b]{0.31\columnwidth}
    \centering
    \includegraphics[width=.8\textwidth]{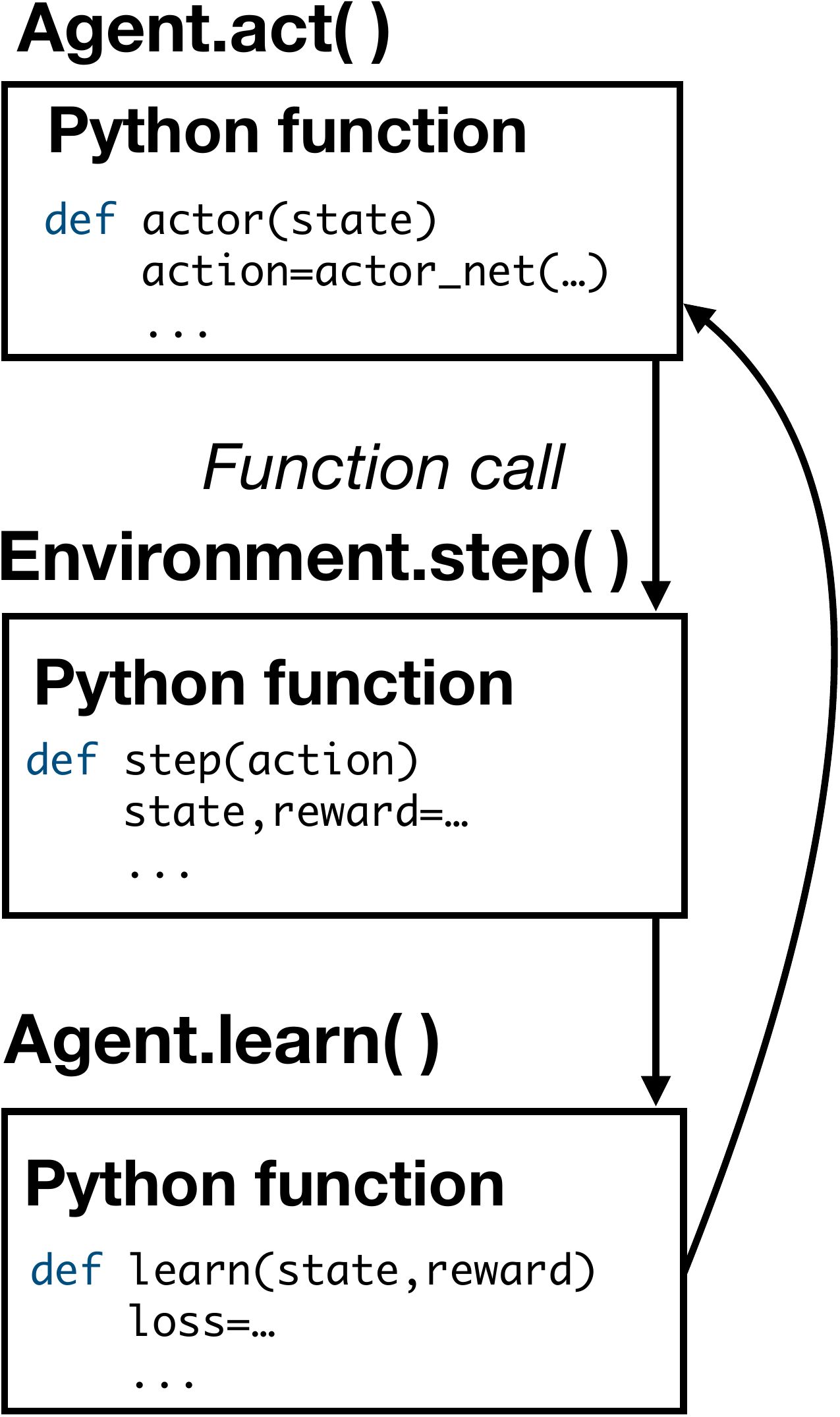}
    \caption{Function-based}\label{fig:ApproachPythonFunc}
  \end{subfigure}
  \begin{subfigure}[b]{0.32\columnwidth}
    \centering
    \includegraphics[width=.8\textwidth]{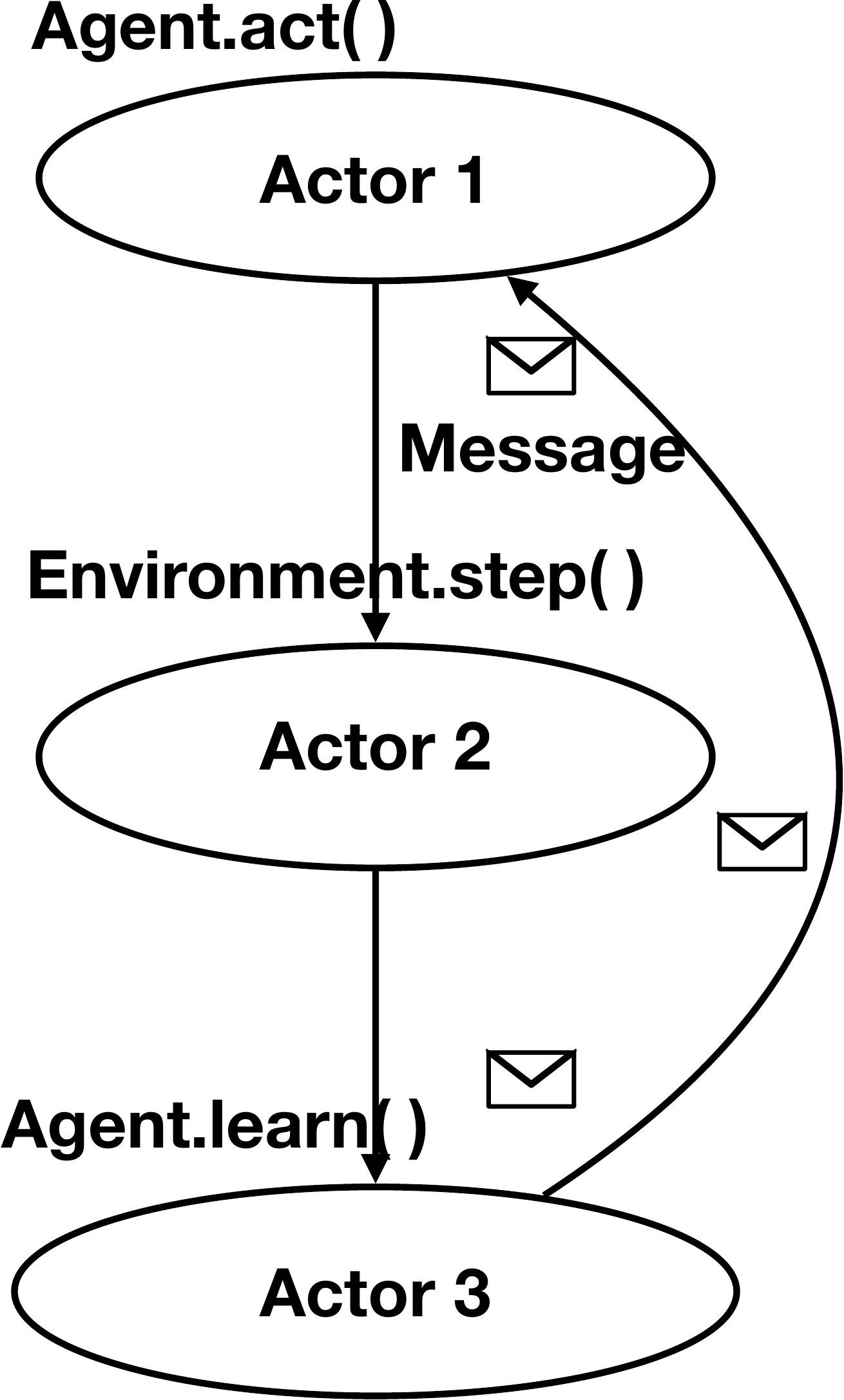}
    \caption{Actor-based}\label{fig:ApproachActor}
  \end{subfigure}
  \begin{subfigure}[b]{0.31\columnwidth}
    \centering
     \includegraphics[width=.8\textwidth]{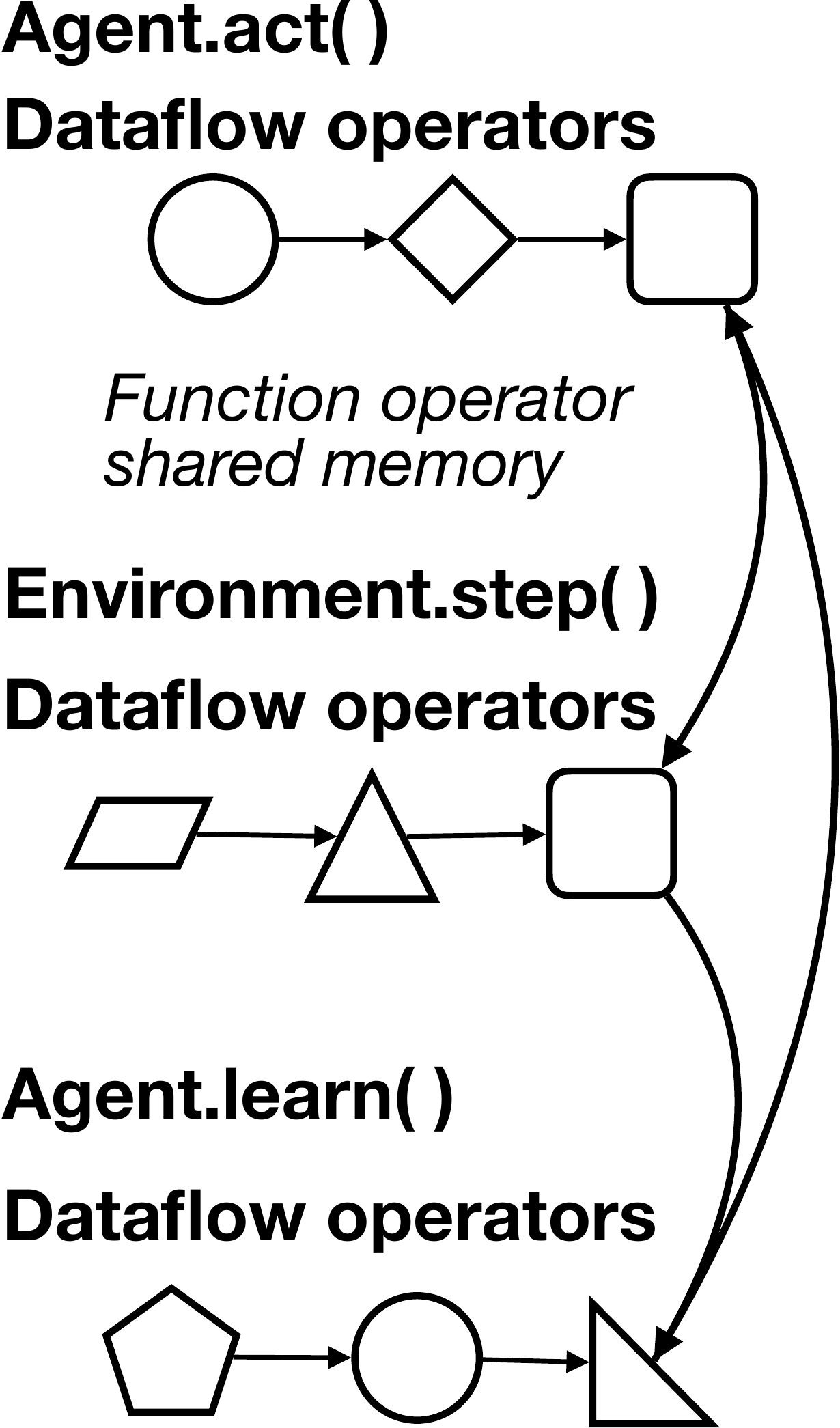}
    \caption{Dataflow-based}\label{fig:ApproachDataflowOperator}
  \end{subfigure}
  \caption{Types of RL system designs}\label{fig:TrainingApproach}
\end{figure}

To understand how existing RL systems support the requirements above, we survey the design space of RL systems. Existing designs can be categorised into three types, \emph{function-based}, \emph{actor-based} and \emph{dataflow-based} (see~\F\ref{fig:TrainingApproach}):

\myparr{(a)~Function-based} RL systems are the most common type. They express RL algorithms typically as Python functions executed directly by workers (see~\F\ref{fig:ApproachPythonFunc}). The RL training loop is implemented using direct function calls.
For example, \emph{Acme}~\cite{DBLP:journals/corr/abs-2006-00979} and \emph{SEED RL}~\cite{DBLP:conf/iclr/EspeholtMSWM20} organise algorithms as actor/learner functions; \emph{RLGraph}~\cite{DBLP:conf/mlsys/SchaarschmidtMF19} uses a component abstraction, and users register Python callbacks to define functionality. Distributed execution is delegated to backend engines, \eg TensorFlow~\cite{TFAgents}, PyTorch~\cite{DBLP:conf/nips/PaszkeGMLBCKLGA19}, Ray~\cite{DBLP:conf/osdi/MoritzNWTLLEYPJ18}.

\myparr{(b)~Actor-based} RL systems execute algorithms through message passing between a set of (programming language) actors deployed on worker nodes~(see~\F\ref{fig:ApproachActor}).
\emph{Ray}~\cite{DBLP:conf/osdi/MoritzNWTLLEYPJ18} defines algorithms as parallel tasks in an actor model. Tasks are distributed among nodes using remote calls. Defining control flow in a fully distributed model, however, is burdensome. To overcome this issue, \emph{RLlib}~\cite{DBLP:conf/icml/LiangLNMFGGJS18} adds logically centralised control on top of Ray. Similarly, \emph{MALib}~\cite{DBLP:journals/corr/abs-2106-07551} offers a high-level agent/evaluator/learner abstraction for population-based MARL algorithms (\eg PSRO~\cite{DBLP:conf/iclr/MullerORTPLHMLH20}) using Ray as a backend.

\myparr{(c)~Dataflow-based} RL systems define algorithms through data-parallel operators, which are mapped to GPU kernels or distributed tasks~(see~\F\ref{fig:ApproachDataflowOperator}). Operators are typically predefined, and users must choose from a fixed set of APIs. For example, \emph{Podracer}~\cite{DBLP:journals/corr/abs-2104-06272} uses JAX~\cite{jax} to compile subroutines as dataflow operators for distributed execution on TPUs. \emph{WarpDrive}~\cite{DBLP:journals/corr/abs-2108-13976} defines dataflow operators as CUDA kernels, and executes the complete RL training loop using GPU thread blocks. \emph{RLlib Flow}~\cite{DBLP:journals/corr/abs-2011-12719} uses distributed dataflow operators, implemented as iterators with message passing.

\label{sec:background:requirements}

\tinyskip

\noindent
\T\ref{tab:design_space} considers how well these approaches satisfy our four requirements from \S\ref{sec:background:dist_rl}:

\mypar{Execution abstraction} Function- and actor-based systems execute RL algorithms directly through implemented (Python) functions and user-defined (programming language) actors, respectively. This prevents systems from applying optimisations how RL algorithms are parallelised or distributed. In contrast, dataflow-based systems execute computation using pre-defined/complied operators~\cite{DBLP:journals/corr/abs-2104-06272, DBLP:journals/corr/abs-2011-12719} or CUDA kernels~\cite{DBLP:journals/corr/abs-2108-13976}, which offers optimisation opportunities.

\mypar{Distribution strategies} Most function-based systems adopt a fixed strategy that distributes actors to parallelise the environment execution (\myc{1}+\myc{2} in \F\ref{fig:MARLTrain}) with only a single learner. Actor-based systems distribute stateful actors and stateless tasks across multiple workers, often using a greedy scheduler without domain-specific planning.

Dataflow-based systems typically hardcode how dataflow operators are assigned to workers. Podracer~\cite{DBLP:journals/corr/abs-2104-06272} provides two strategies: \emph{Anakin} co-hosts an environment and an agent on each TPU core; \emph{Sebulba} distributes the environment, learners and actors on different TPU cores; and RLlib Flow~\cite{DBLP:journals/corr/abs-2011-12719} shards dataflow operators across distributed Ray actors.

\mypar{Acceleration support} Most RL systems only accelerate DNN policy inference and training~(\myc{1}+\myc{3}). Some dataflow-based systems (\eg Podracer~\cite{DBLP:journals/corr/abs-2104-06272} and WarpDrive~\cite{DBLP:journals/corr/abs-2108-13976}) also accelerate other parts of training, requiring custom dataflow implementations: \eg Podracer can accelerate the environment execution~(\myc{2}) on TPU cores; WarpDrive executes the entire training loop~(\myc{1}--\myc{3}) on a single GPU using CUDA. 

\mypar{Algorithm abstraction} Function-based RL systems make it easy to provide intuitive actor/learner/env APIs. Actor-based RL systems often provide lower-level harder-to-use APIs for distributed components (\eg Ray's \texttt{get}/\texttt{wait}/\texttt{remote}~\cite{DBLP:conf/osdi/MoritzNWTLLEYPJ18}); higher-level libraries (\eg RLlib's \texttt{PolicyOptimizer} API~\cite{DBLP:conf/icml/LiangLNMFGGJS18}) try to bridge this gap. Dataflow-based systems come with fixed dataflow operators, requiring users to rewrite algorithms. For example, JAX~\cite{jax} provides the \texttt{vmap} operator for vectorisation and \texttt{pmap} for single-program multiple-data~(SPMD) parallelism. 

\tinyskip

\noindent
In summary, there is an opportunity to design a new RL system that combines the usability of function-based  systems with the acceleration potential of dataflow-based systems. Such a design requires a new execution abstraction that retains the flexibility to apply different acceleration and distribution strategies on top of an RL algorithm specification.


\section{\sys API}
\label{sec:api}

\begin{lstlisting}[language={Python}, xleftmargin=2em,,framexleftmargin=1.5em, linewidth=0.48\textwidth, basicstyle=\linespread{0.7}\scriptsize\ttfamily, numbersep = 4pt, stringstyle=\color{purple}, keywordstyle=\color{black}\bfseries, commentstyle=\color{olive}, float=!h, label=alg:mappo, caption=MAPPO algorithm in \sys, tabsize=2, numbers=left, breaklines=true, breakatwhitespace=false, numberblanklines=false, emph={MAPPOAgent, Agent, MAPPOActor, Actor, Learner, MAPPOLearner, Trainer, MAPPOTrainer, MAPPO, MAPPOActorNet, MAPPOCriticNet, MAPPONetTrain,MultiActorEnvSingleLearner_DP, DistributionPolicy,set\_boundary, set\_fragment_number, fuse\_fragment, set\_replicate\_list, mappo\_algorithm\_config, learner\_params, env\_params, mappo\_deployment\_config, AthenaRL, env\_step, env\_reset, co\_locate, mapping}, emphstyle={\color{blue}\bfseries}, escapeinside={(*@}{@*)}]
class MAPPOAgent(Agent): (*@\label{mappo:agent}@*)
 def act(self,state):
   return self.actors.act(state)
 def learn(self,sample):
   return self.learner.learn(sample)
  	
class MAPPOActor(Actor) (*@\label{mappo:actor}@*)
 def act(state):
   action = self.actor_net(state) (*@\label{mappo:action}@*)
   #@MSRL.fragment(type='Action',ops=['AllGather'],data=[action])(*@\label{mappo:anno_action}@*)
   reward,new_state = (*@\bfseries\color{blue}MSRL.env\_step(action)@*) (*@\label{mappo:step}@*)
   #@MSRL.fragment(type='Step',ops=['AllGather'],data=[reward,new_state])(*@\label{mappo:anno_envstep}@*)
   return reward,new_state
    
class MAPPOLearner(Learner): (*@\label{mappo:learner}@*)
 def learn(sample):
   action,reward,state,next_state = sample (*@\label{mappo:sample}@*)
   last_pred = self.critic_net(next_state)
   pred = self.critic_net(state)
   r = discounted_reward(reward,last_pred,self.gamma)
   adv = gae(reward,next_state,pred,last_pred,self.gamma)
   for i in range(self.iter):
     loss += self.mappo_net_train(action,state,adv,r)(*@\label{mappo:learn_end}@*)
   return loss / self.iter
 	  	
class MAPPOTrainer(Trainer): (*@\label{mappo:trainer}@*)
 def train(self,episode):
   for i in range(episode):(*@\label{mappo:enter}@*)
     state = (*@\bfseries \color{blue} MSRL.env\_reset() @*) (*@\label{mappo:reset}@*)
     #@MSRL.fragment(type='Reset',ops=['AllGather'],data=[state])(*@\label{mappo:anno_reset}@*)	   
     for j in range(self.duration):
       reward,new_state = (*@\bfseries\color{blue} MSRL.agent\_act(state)@*) (*@\label{mappo:act}@*)
       (*@\bfseries\color{blue} MSRL.replay\_buffer\_insert(reward,new\_state)@*) (*@\label{mappo:buff}@*)
     sample = (*@\bfseries\color{blue} MSRL.replay\_buffer\_sample()@*)
     #@MSRL.fragment(type='Buffer',ops=['AllGather'],data=[sample]) (*@\label{mappo:anno_buff}@*)
     loss = (*@\bfseries\color{blue} MSRL.agent\_learn(sample)@*)(*@\label{mappo:learn}@*)
     #@MSRL.fragment(type='Learner',ops=['AllGather'], data=[actor_net.get_trainable_params()])(*@\label{mappo:anno_par}@*) 	  	

mappo_algorithm_config = { (*@\label{mappo:config:alg}@*)
 'agent':  {'num':4,'type':MAPPOAgent,
            'actor':MAPPOActor,'learner':MAPPOLearner},
 'actor':  {'num':1,'type':MAPPOActor, 
            'policy':MAPPOActorNet,'env':True},
 'learner':{'num':1,'type':MAPPOLearner, 
            'policy':[MAPPOCriticNet,MAPPONetTrain], 
            'params':{'gamma':0.9}},
 'env':{'type':MPE,'num':32,'params':{'name':'MPE'}}} (*@\label{mappo:config:alg:end}@*)

mappo_deployment_config = { (*@\label{mappo:config:deploy}@*)
 'workers':[198.168.152.19, 198.168.152.20, (*@$[\ldots]$@*), 
 'GPUs_per_worker':2},
 'distribution_policy':'Single_learner_coarse'}(*@\label{mappo:config:end}@*) 
\end{lstlisting}

\noindent
Next, we introduce \sys{}'s API and give an example of how it can implement the multi-agent PPO~(MAPPO) algorithm~\cite{mappo2021}. The API separates the RL algorithm's logic from deployment and execution considerations.

A user expresses an RL algorithm through familiar concepts, such as agents, actors, learners, trainers and environments. An \emph{agent} consists of actors and learners: \emph{actors} collect data from the \emph{environment}, and \emph{learners} manage the training policy. A \emph{trainer} provides the training loop logic, specifying the interaction between actors and learners for each agent.

\sys offers two types of APIs: (i)~a \emph{component API} defines the behaviour of each algorithmic component, defined through abstract classes. For example, \code{Actor.act()} specifies the interaction between actors and environments; \code{Learner.learn()} implements the training logic of the neural network; a training loop is defined through \code{Trainer.train()}; and (ii)~an \emph{interaction API} allows components to interact. For example, an actor submits the collected training data using \code{MSRL.replay\_buffer\_insert()}; a learner samples from the replay buffer using \code{MSRL.replay\_buffer\_sample()}.

\sys determines how the algorithm is deployed through two configurations: (1)~an \emph{algorithm configuration} defines the logical components (\ie agents, actors, learners etc.) and their hyper-parameters; and (2)~a \emph{deployment configuration} refers to a \emph{distribution policy} and the devices.

\mypar{Example} \A\ref{alg:mappo} sketches the MAPPO implementation. Its policies and DNNs are omitted for simplicity. \code{MAPPOAgent}~(line~\ref{mappo:agent}) defines the agent behaviour: it interacts with the environment through \code{MAPPOActor}~(line~\ref{mappo:actor}), and performs the policy training with \code{MAPPOLearner}~(line~\ref{mappo:learner}). It then collects experiences from the environment with a given policy~(lines~\ref{mappo:action}--\ref{mappo:step}), which is used to optimise the DNN model~(lines~\ref{mappo:sample}--\ref{mappo:learn_end}). To define behaviour, \code{MAPPOAgent}, \code{MAPPOActor} and \code{MAPPOLearner} inherit from base classes, overriding abstract methods with the algorithm's logic. 

\code{MAPPOTrainer} defines the training loop~(line~\ref{mappo:trainer}) and uses \sys{}'s interaction API. After entering the loop~(line~\ref{mappo:enter}), it resets the environment using \code{MSRL.env\_reset()}. It then uses \code{MSRL.agent\_act()} to collect the experience (line~\ref{mappo:act}) from \code{MAPPOActor}, and saves it in the replay buffer~(line~\ref{mappo:buff}). Finally, the trainer calls \code{MSRL.agent\_learn()} to execute the \code{learn()} function from \code{MAPPOLearner}.

\mypar{Partition annotations} Despite defining an RL algorithm in a classical way with actors, learners and trainers, \sys must be able to decompose the algorithm and distribute computation across devices. Since auto-distributing code is a hard problem~\cite{legion, dandelion, 10.1145/3437801.3441587}, \sys relies on user \emph{annotations} in the code to help identify suitable partitioning boundaries.

The \emph{partition annotations} in lines~\ref{mappo:anno_action},~\ref{mappo:anno_envstep},~\ref{mappo:anno_reset},~\ref{mappo:anno_buff},~\ref{mappo:anno_par} denote possible boundaries in the algorithm for parallel and distributed execution. Each annotation specifies the type of code fragment and what dependent data must be transferred at the boundary when computation is distributed. Code between two consecutive partition annotations thus becomes a self-contained \emph{fragment}, which can be assigned to a device.

As an example, consider the fragment defined by the annotations in lines~\ref{mappo:anno_action} and~\ref{mappo:anno_envstep}: the first annotation stipulates that an \code{action} can be received from the previous fragment before interacting with the environment; the second annotation sends the \code{reward} and \code{new\_state} to the next fragment on another device. As we explain in the next section, this enables \sys to distribute the environment simulation across devices.

\myparr{\textnormal{The} algorithm configuration \textnormal{(lines~\ref{mappo:config:alg}--\ref{mappo:config:alg:end})}} is a Python dictionary that defines how to instantiate the algorithmic components (lines~\ref{mappo:config:alg}). It specifies their number, type and hyper-parameters, which \sys requires for execution~(see~\S\ref{sec:design}). In this example, the training uses 4~agents, each with 1~actor and 1~learner; and each actor interacts with 32~environments.

\myparr{\textnormal{The} deployment configuration \textnormal{(lines~\ref{mappo:config:deploy}--\ref{mappo:config:end})}} specifies how the computation is executed through a \emph{distribution policy}. With the help of the partition annotations, the distribution policy governs the partitioning of the algorithmic components and their distribution. We discuss partitioning and distribution policies in~\S\ref{sec:dataflow_model} and \S\ref{sec:distribution}, respectively.


\section{Translation to Fragmented Dataflow Graphs}
\label{sec:dataflow_model}

We now describe how \sys uses fragmented dataflow graphs to decouple the parallelisation and distribution of RL algorithms from their implementation.

\begin{figure}[tb]
  \centering
  \includegraphics[width=0.9\columnwidth]{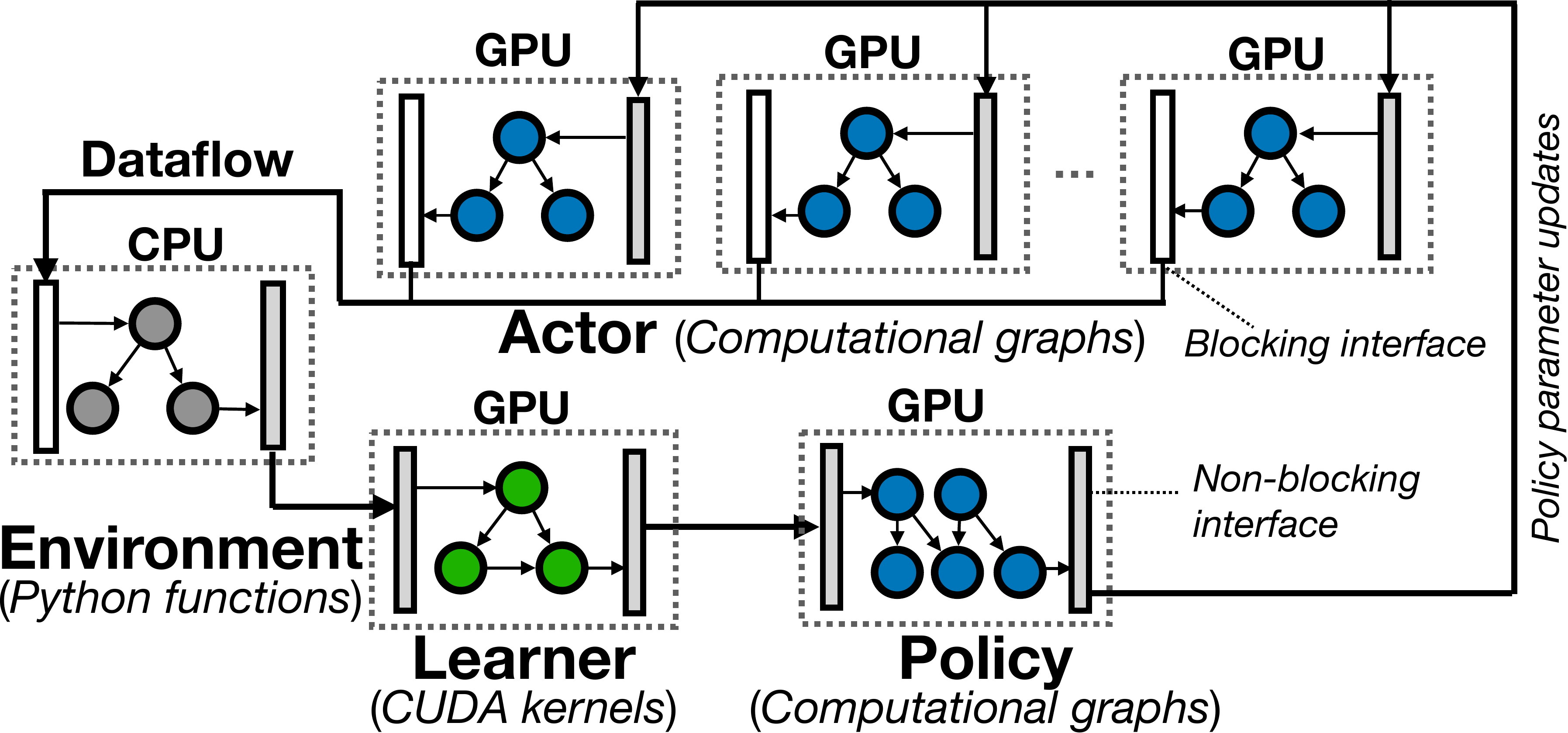}
  \caption{Translating functions to fragmented dataflow graphs}\label{fig:dataflow}
\end{figure}

\subsection{Fragmented dataflow graphs}

\noindent
\sys must partition an RL algorithm so that it can be distributed across workers. Since it is challenging to parallelise and distribute the algorithm implementation directly, \sys introduces a new execution abstraction by representing the algorithm as a \emph{fragmented dataflow graph}~(FDG): each node in the FDG is a potentially data-parallel fragment of code, which can be deployed on workers with GPUs and CPUs. FDGs thus allow \sys to execute the RL computation on different devices depending on the allocation of fragments. \sys distributes fragments to devices on remote workers, which then exchange data over the network; or fragments may be co-located on a single worker and use shared memory communication.

\F\ref{fig:dataflow} gives a high-level overview of an FDG, which consists of multiple dataflow \emph{fragments} (dashed square nodes). A fragment encompasses the dataflow graph of the corresponding code with two interfaces: an \emph{entry} and an \emph{exit} interface. These interfaces implement the functionality for sharing data with other fragments: if fragments are located on the same worker, they can directly share data structures; if they belong to different workers, they can use network communication operators. Depending on the communication method, the interface may be blocking (\eg when using shared data structures) or non-blocking (\eg for asynchronous communication). 

Each fragment must be allocated to a device. Depending on how a fragment's code is implemented, fragments require specific hardware resources for execution. For example, the \emph{actor} and the \emph{policy} fragments in \F\ref{fig:dataflow} are implemented using the data-parallel operators of a DL engine, and thus require a GPU for execution; the \emph{environment} fragment has a native Python implementation that executes on a CPU. 

\sys parallelises the execution of fragment code by creating multiple instances of it and assigning each instance to a separate device. All such replicated fragments share the same interface, allowing \sys to use the communication operation from the partition annotation to synchronise data~(see~\S\ref{sec:api}). For example, if the fragment denoted by the partition annotations in lines~\ref{mappo:anno_buff} and~\ref{mappo:anno_par} in \A\ref{alg:mappo} is replicated, the instances synchronise using the \code{AllGather} method.

\subsection{Deployment using fragment dataflow graphs}
\label{sec:dataflow:fragment_deployment}

\begin{figure}[tb]
  \centering
  \includegraphics[width=.7\columnwidth]{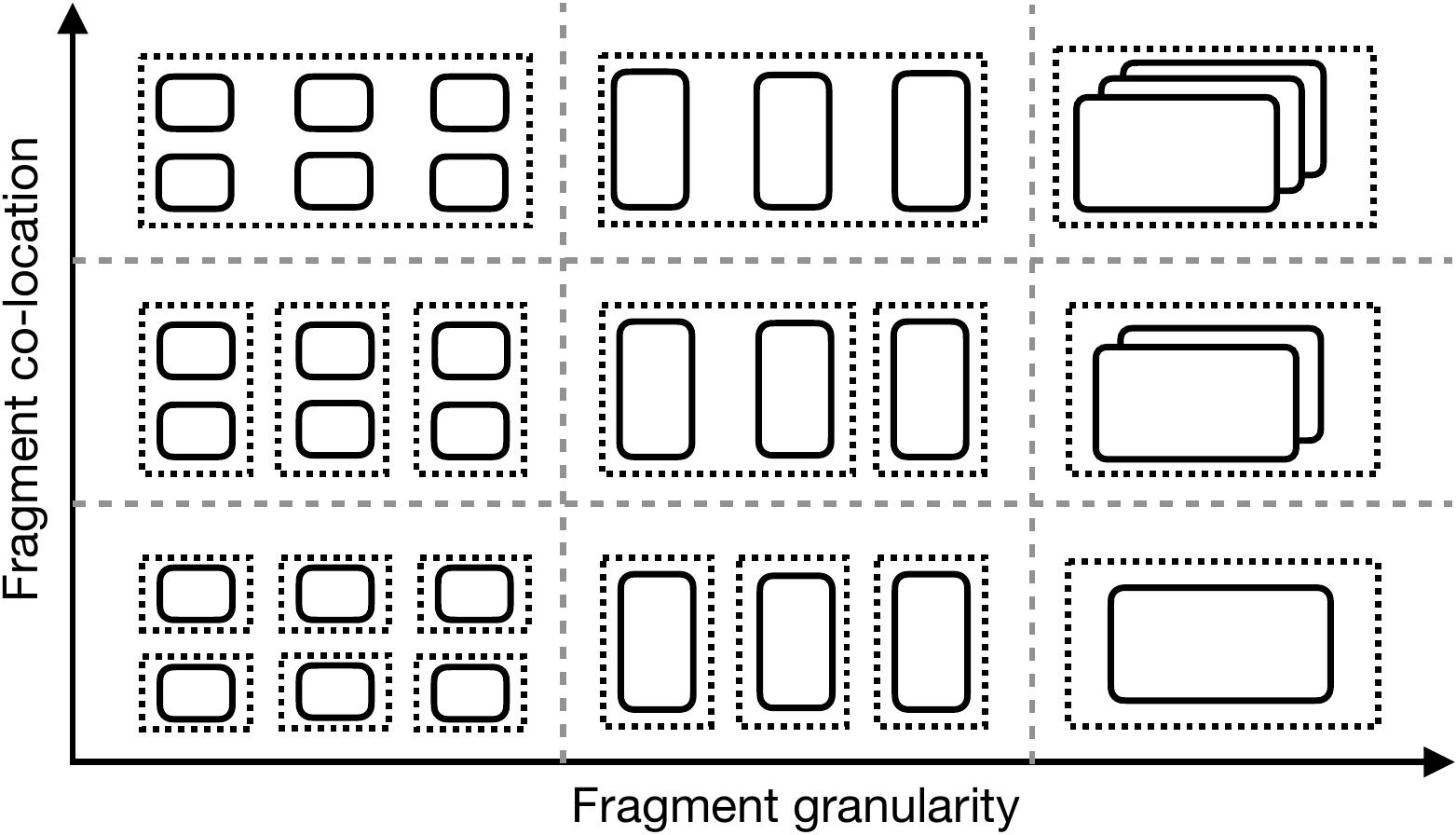}
  \caption{Fragment granularity vs. fragment co-location}\label{fig:fragmemt_spectrum}
\end{figure}
 
When deploying an RL algorithm as an FDG, two dimensions impact execution performance~(see~\F\ref{fig:fragmemt_spectrum}):

\myparr{(1)~Fragment granularity} refers to a fragment's code size, which affects device utilisation. A small fragment may underutilise a GPU; a large one may exhaust GPU memory.

The fragment granularity also determines the ratio between computation and communication. The frequency and amount of data exchanged during synchronisation between fragments is a major source of overhead, often limiting scalability. In general, coarser fragments require less synchronisation, which reduces communication overhead, but they diminish opportunities for parallelism. For example, multiple fragments may exchange policy parameters (\eg weights) or the replay buffer (\ie trajectories) frequently at each step; alternatively, they may batch data from multiple steps and communicate at each episode to increase network efficiency.

FDGs can thus subsume the existing execution strategies of current RL systems. Based on the partition annotations, an FDG may fuse an actor and environment into one CPU-based fragment, and a learner into a GPU-based fragment (\eg as proposed by Acme~\cite{DBLP:journals/corr/abs-2006-00979}); alternatively, it can create a coarser-grained GPU fragment by moving the DNN from the CPU fragment to the learner, which accelerates policy inference (as done by SEED RL~\cite{DBLP:conf/iclr/EspeholtMSWM20}); at the coarsest granularity, an even larger fragment may contain the actor, learner, DNNs and environment, which accelerates the whole training loop (as done by WarpDrive~\cite{DBLP:journals/corr/abs-2108-13976} or Anakin~\cite{DBLP:journals/corr/abs-2104-06272}).

\myparr{(2)~Fragment co-location} refers to the mapping of fragments to devices on the same worker. Co-locating fragments avoids network communication (Ethernet or InfiniBand) for synchronisation. Instead, it uses intra-node communication (NVLink or PCIe), which reduces latency and increases throughput. Whether co-location is feasible depends on the available resources (\eg the number of GPUs per node) and fragment constraints (\eg the resource needs of fragments).

\tinyskip

\noindent
Choosing the right trade-off between fragment granularity and co-location is key to achieving good performance. The FDG abstraction allows \sys to make different trade-offs using a set of distribution policies. Next, we describe how \sys transforms an RL algorithm into an FDG.

\subsection{Building fragmented dataflow graphs}

\sys generates an FDG by partitioning the algorithm's dataflow graph. Based on the user-specified partition annotations (\S\ref{sec:api}), it splits the dataflow graph into fragments, which then exchange the data specified by the annotations. 

\setcounter{algorithm}{1}
\begin{algorithm}[tb]
\caption{Fragmented dataflow graph generation}\label{alg:build_fdg}
{\fontsize{9}{9}\selectfont
\begin{algorithmic}[1]

\Function{generate\_fdg}{algorithm}
\State $\textit{FDG} \gets \{\}$
\State $\textit{DFG} \gets \textit{generate\_dfg(algorithm)}$ \label{alg:dfg}
\State $\textit{boundary} \gets \textit{annotation\_parser(annotations)}$\label{alg:anno_parser}
\State $\textit{common\_node} \gets \textit{label\_common\_node(DFG, boundary)}$\label{alg:common}
\For {$\textit{node}$ in $\textit{common\_node}$}
\State $\textit{sub\_graph} \gets \textit{graph\_traversal(node, DFG)}$ \label{alg:travel}
\State $\textit{fragment} \gets \textit{generate\_interface(sub\_graph, boundary})$\label{alg:interface}
\State $\textit{FDG} \gets FDG \cup \textit{fragment}$
\EndFor

\State \Return FDG
\EndFunction
\end{algorithmic}
}
\end{algorithm}

\begin{figure}[tb]
  \centering
  \begin{subfigure}[b]{0.12\textwidth}
    \centering
    \includegraphics[width=0.9\textwidth]{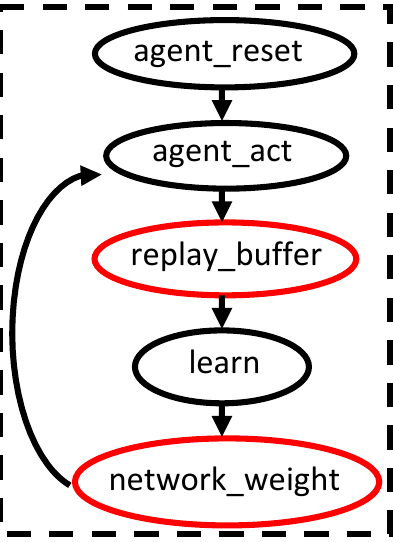}
    \caption{Dataflow graph}\label{fig:ppo_data_flow}
  \end{subfigure}
  \begin{subfigure}[b]{0.3\textwidth}
    \centering
    \includegraphics[width=0.99\textwidth]{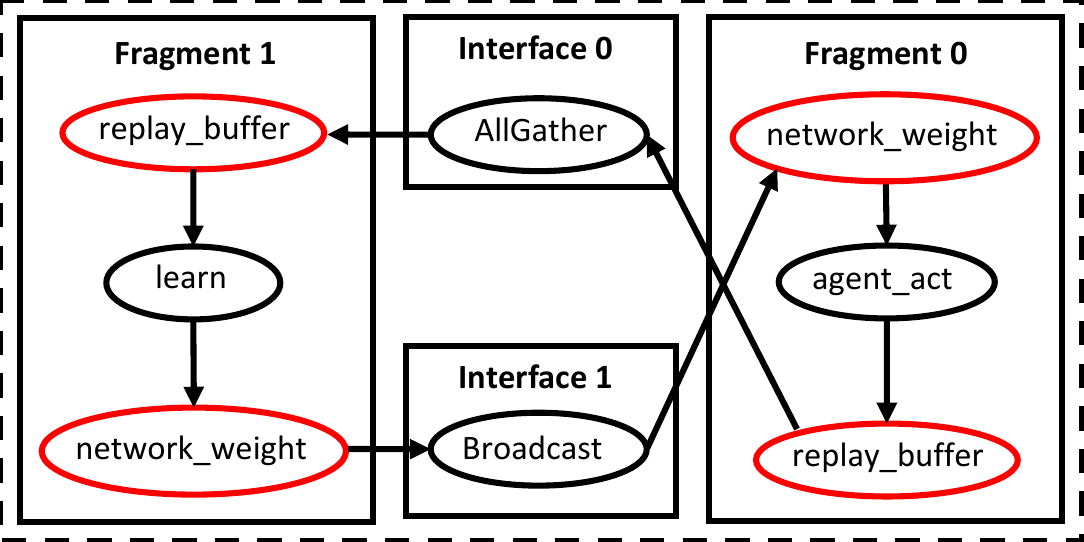}
    \caption{Two fragments}\label{fig:ppo_fragment}
  \end{subfigure}
  \caption{Example of FDG construction for MAPPO}\label{fig:fdg-algorithm}
\end{figure}

\A\ref{alg:build_fdg} shows the algorithm to generate an FDG. First, it statically analyses the RL algorithm to construct a dataflow graph, and then it partitions the graph into fragments. 

Recall that the partition annotations denote the boundaries between code segments and describe the data that must be transferred when the code is partitioned. Correspondingly, in the partitioned fragments, the nodes that represent the data specified by the partitioned annotation must be at the boundary of fragments. We refer to these nodes as \emph{common} nodes in the dataflow graph. \sys creates fragments by partitioning the graph at the common nodes.

Consider the partitioning of the MAPPO algorithms from \A\ref{alg:mappo} with the two annotations in lines~\ref{mappo:anno_buff} and~\ref{mappo:anno_par}. \F\ref{fig:ppo_data_flow} shows its simplified dataflow graph, with the data described by the two annotations highlighted in red. By splitting the graph at the common nodes, it can be partitioned into two fragments (see~\F\ref{fig:ppo_fragment}). In each fragment, the common nodes (circled in red) are connected to interface implementations, which are generated for a given communication method.  

To perform the partitioning, \A\ref{alg:build_fdg} first parses the partition annotations to record the communication method and corresponding data for each fragment boundary~(line~\ref{alg:anno_parser}) and uses this information to label the common nodes~(line~\ref{alg:common}). It then traverses the dataflow graph from each common node~(line~\ref{alg:travel}). The traversal terminates when it reaches another common node or a leaf node, and the constructed subgraph forms a new fragment. The algorithm also duplicates the common nodes in the original dataflow graph and fragment graph. Since the subgraph between two common nodes can only belong to one fragment, the algorithm removes the subgraph from the original dataflow graph after constructing the fragment, removing it from the following search. Finally, it adds the implementation of the communication interfaces and connects the common nodes with the interfaces~(line~\ref{alg:interface}).

If the user does not provide any partition annotations, \sys, by default, partitions the algorithm along the algorithmic components, \ie actors, learners, environments etc. Each component's input and output values are the common nodes in the dataflow graph. The communication method is chosen by analysing the dataflow between the components. 



\section{\sys Architecture}
\label{sec:design}

\begin{figure}[tb]
  \centering
  \includegraphics[width=.48\textwidth]{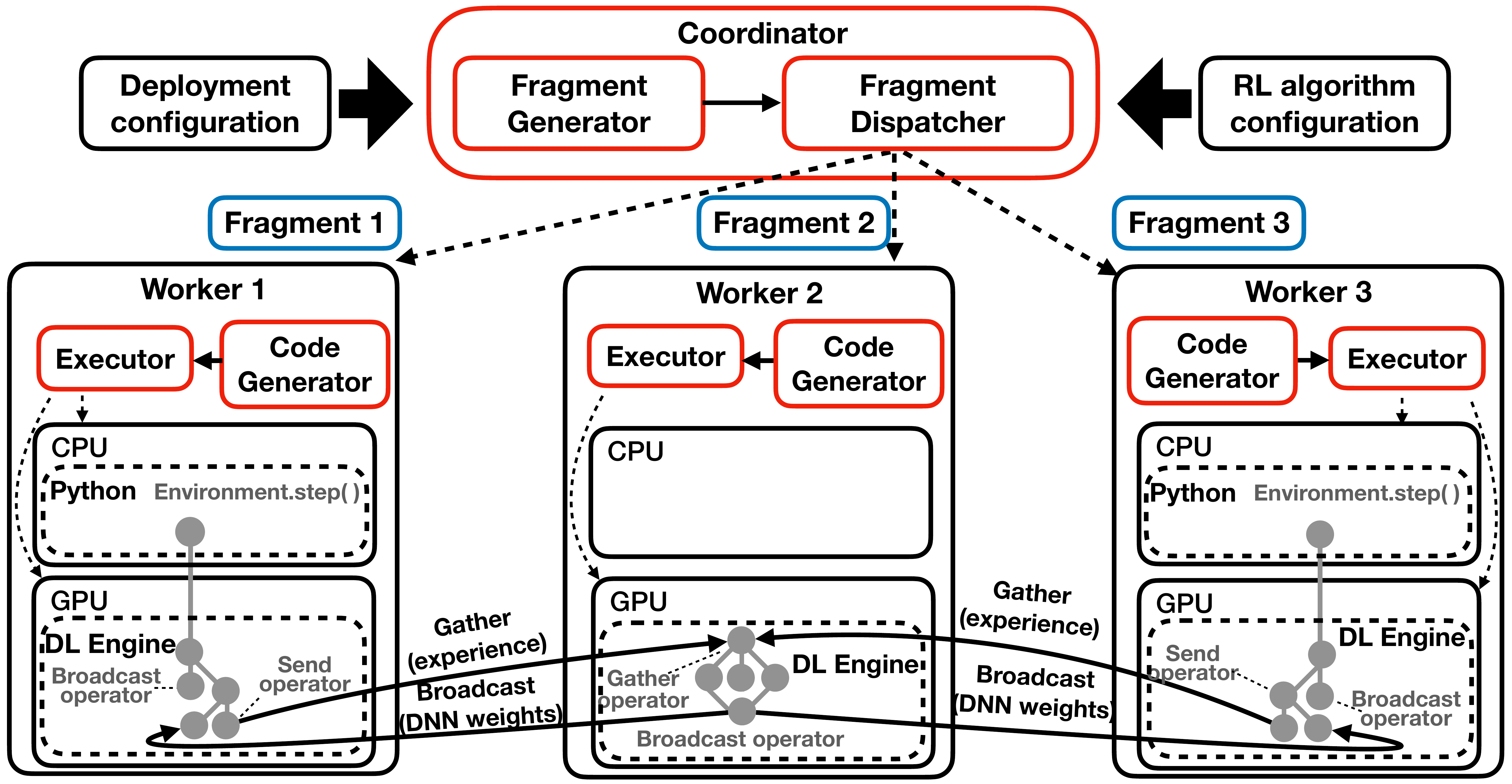}
  \caption{Overview of the \sys architecture\label{fig:Framework}}
\end{figure}

We describe \sys{}'s architecture, which follows a coordinator/worker design (see~\F\ref{fig:Framework}): (1)~a user submits the RL algorithm implementation to a coordinator; (2)~the coordinator generates the FDG and dispatches fragments to workers; and (3)~workers generate executable code for the received fragments, which can be run by \eg a DL engine.

\begin{table*}[t!]
  \small\selectfont
  \begin{tabular}{p{4.2cm}@{\hspace{4pt}}p{5.4cm}p{7.2cm}}
    \toprule
    \textbf{Distribution policy} & \textbf{Deployment} & \textbf{Description}\\

    \midrule
    
    \textbf{[DP-A]: Single learner/coarse} \newline \newline
    replicate: (\emph{actor}, \emph{env}) \newline split: (\emph{learner})\label{fig:distributionMultiActor} \newline \newline
    \eg Acme~\cite{DBLP:journals/corr/abs-2006-00979}, Sebulba~\cite{DBLP:journals/corr/abs-2104-06272} &
    \includegraphics[align=t, width=\linewidth]{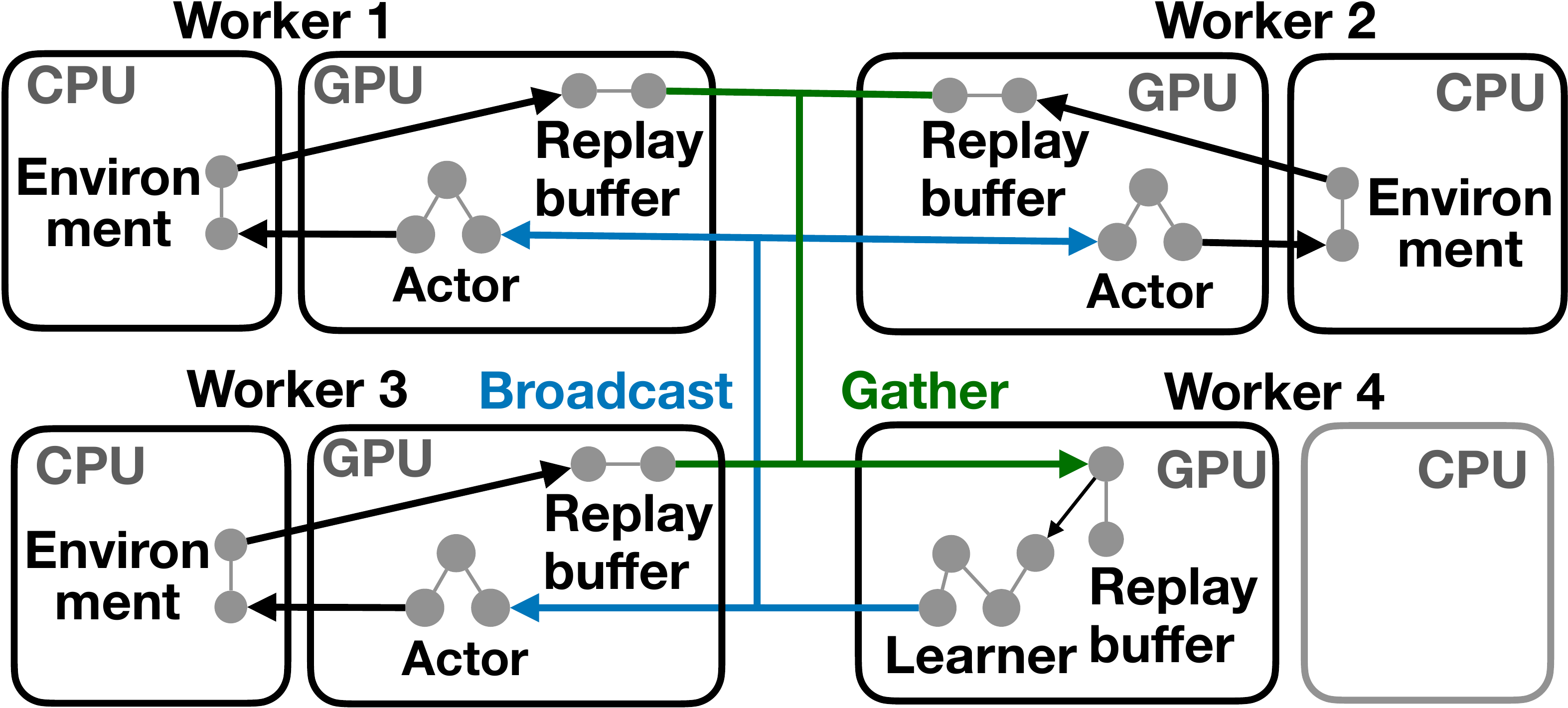} & DP-A replicates the actor and environment fragments: W1--W3 co-locate 1~GPU fragment with an actor for DNN policy inference and 1~CPU fragment for the environment execution. A single GPU fragment with a learner performs policy training (W4), gathering batched training data, training the policy and broadcasting updates.\\

    \midrule

    \textbf{[DP-B] -- Single learner/fine} \newline \newline
    replicate: fused \emph{actor/env} \newline split: \emph{learner}\label{fig:distributionMultiActorCPU}  \newline \newline
    \eg SEED RL~\cite{DBLP:conf/iclr/EspeholtMSWM20} & 
    \includegraphics[align=t, width=\linewidth]{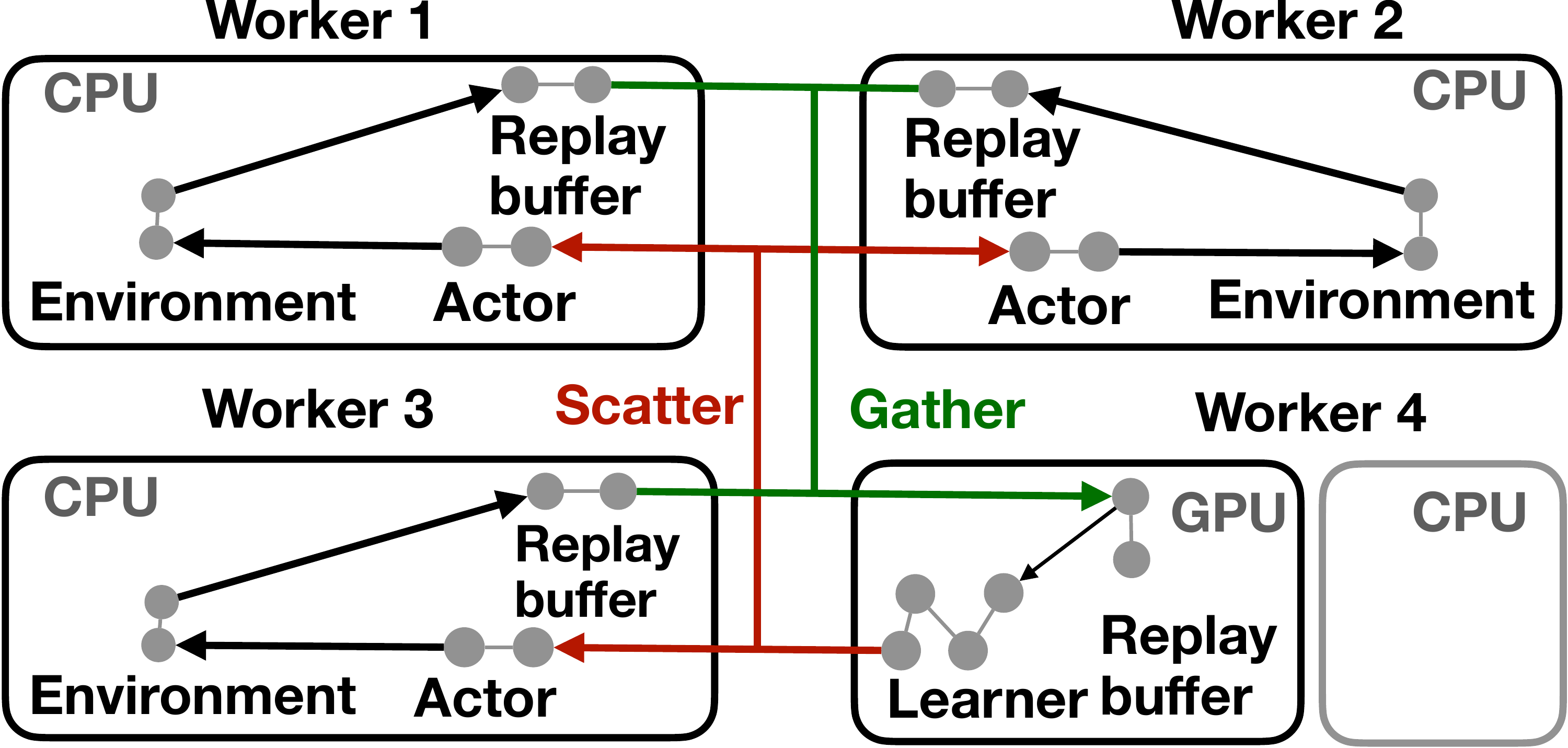} & DP-B fuses the actor and environment into 1~fragment (W1--W3) but handles policy inference at the learner (W4), \ie actors do not contain DNNs. W4 executes policy inference and training in 1~GPU fragment; W1--3 only have CPU fragments. W4 scatters actions to W1--W3 and gathers data for policy training.\\

    \midrule
    
    \textbf{[DP-C] -- Multiple learners} \newline \newline
    replicate: fused \emph{actor/learner}, \emph{env}\label{fig:distributionAgent} &
    \includegraphics[align=t, width=\linewidth]{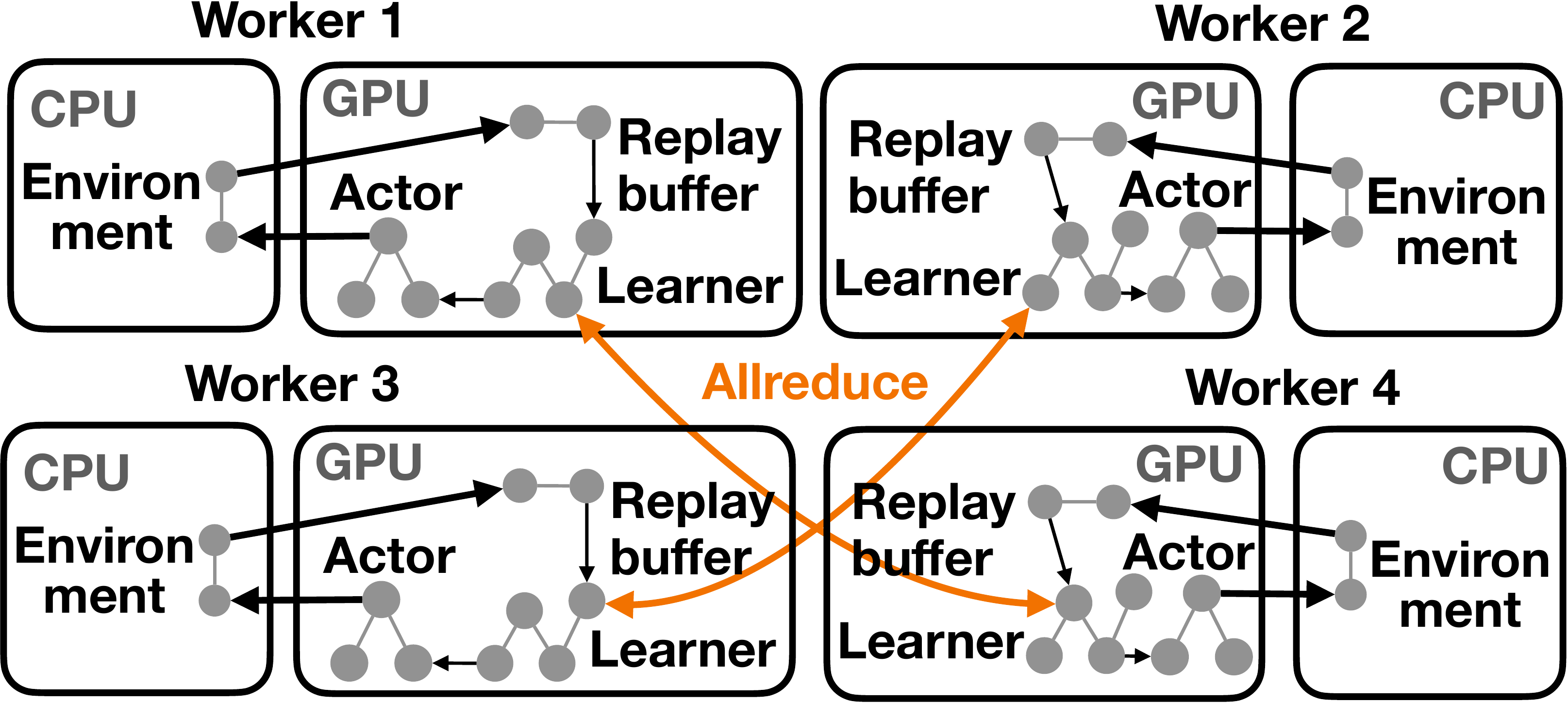} & DP-C performs data-parallel training with multiple learners, supporting fully decentralised MARL training~\cite{DBLP:conf/icml/ZhangYL0B18, DBLP:conf/icml/0003KWGL20, DBLP:conf/nips/QuMXQSX19, DBLP:conf/icml/ZimmerGSW21}. DP-C co-locates 2~fragments: a GPU fragment that fuses the actor and learner, accelerating policy inference, training and replay buffer management, and a CPU fragment for environment execution.\\

    \midrule
    
    \textbf{[DP-D] -- GPU only} \newline \newline
    replicate: \mbox{fused \emph{actor/learner/env}}\label{fig:distributionAgentGPU} &
    \includegraphics[align=t, width=\linewidth]{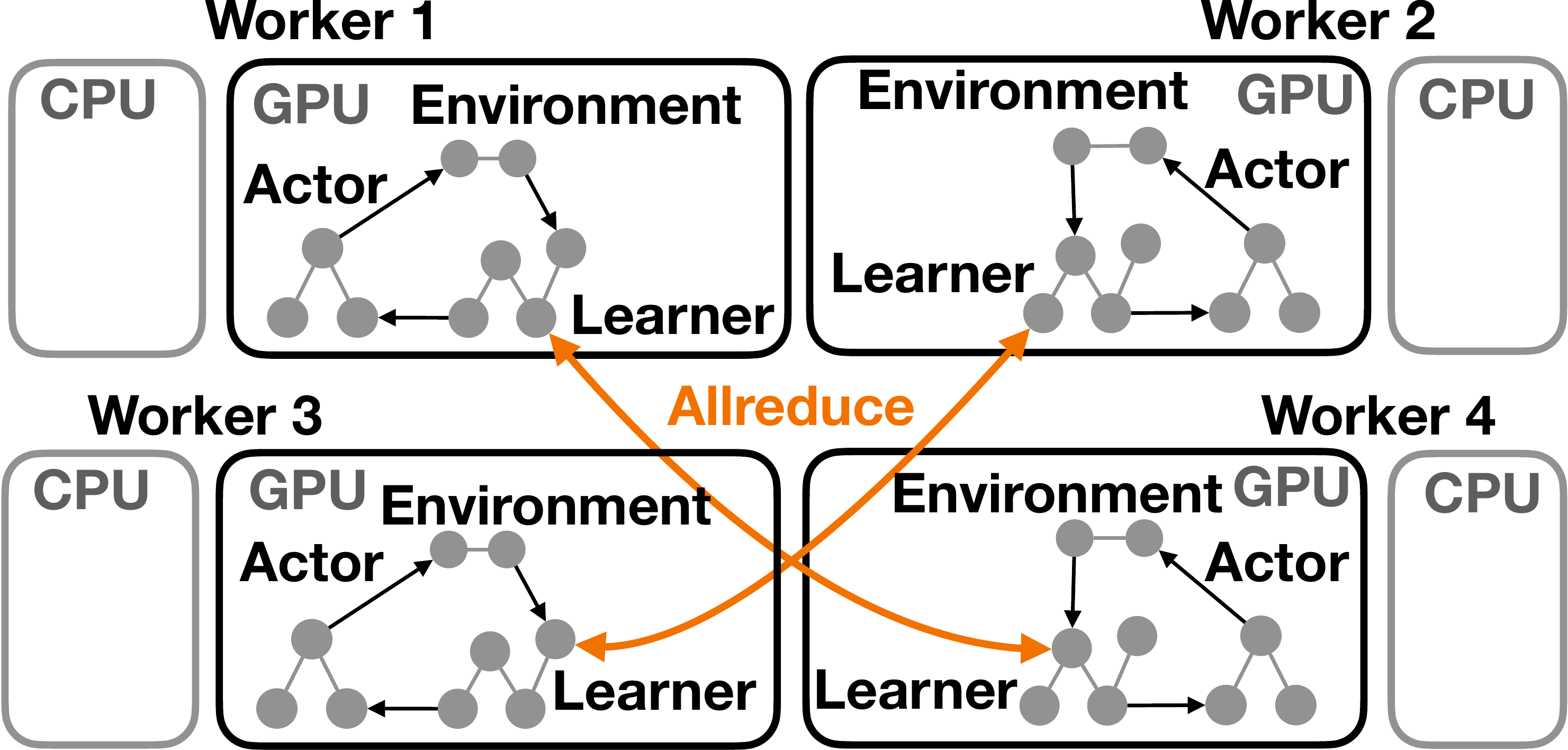} & DP-D fuses the training loop into 1~GPU fragment. To enable communication among GPU fragments, DP-D uses \code{Allreduce} operators compiled into the computational graph with NCCL2~\cite{nccl}. DP-D is a distributed implementation of the single-node systems (\eg WarpDrive~\cite{DBLP:journals/corr/abs-2108-13976}).\\
    
    \midrule
    
    \textbf{[DP-E] -- Environments} \newline \newline
    replicate: fused \emph{actor/learner} \newline split: \emph{env}\label{fig:distributionEnvWorker} \newline \newline
    \eg MALib~\cite{DBLP:journals/corr/abs-2106-07551} &
    \includegraphics[align=t, width=\linewidth]{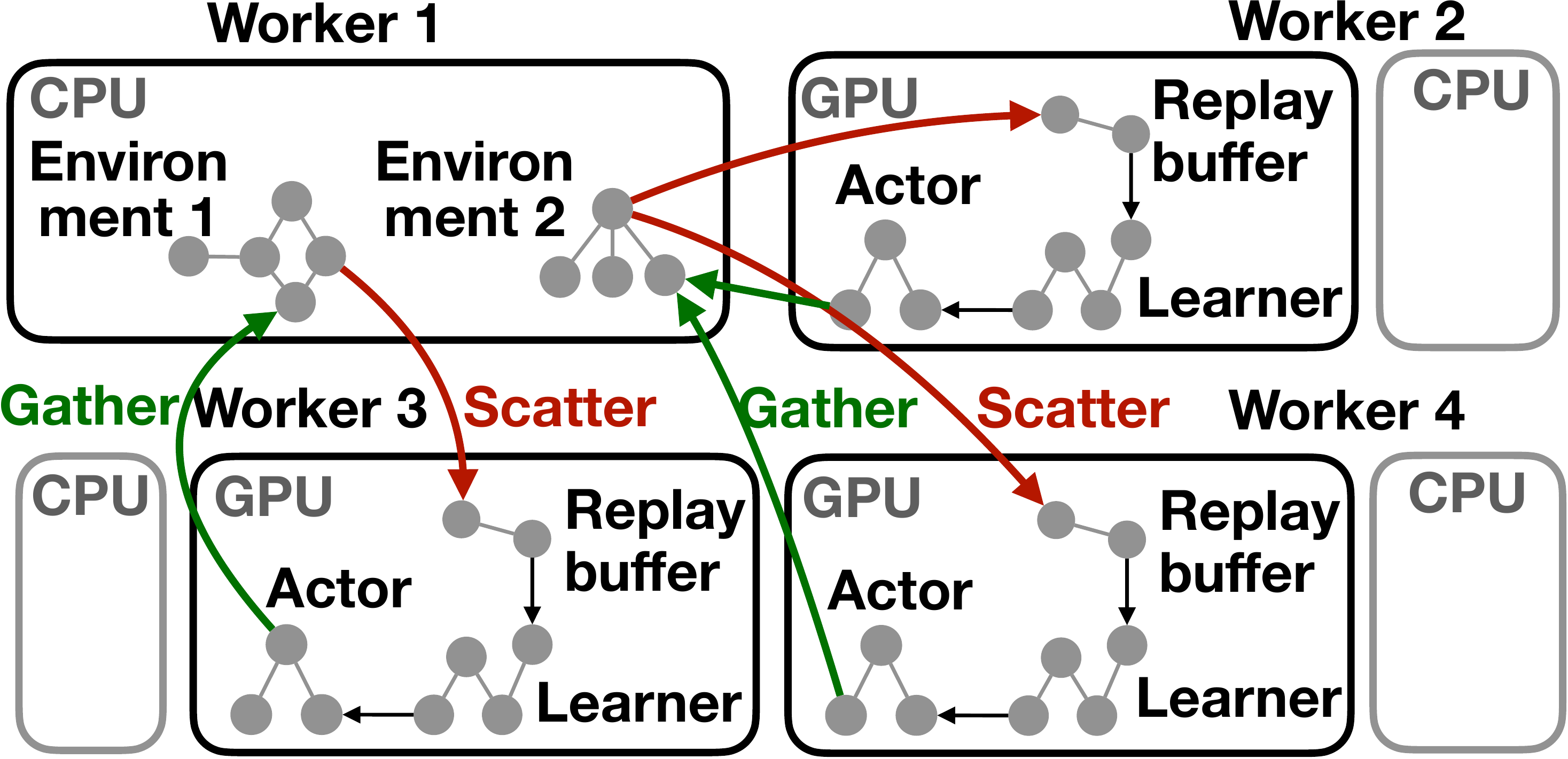} & DP-E has a dedicated worker for environment execution. W1 has CPU fragments to execute environment instances on multiple CPU cores; W2--W4 fuse the actor and learner to accelerate policy inference and training. W1 gathers the inferred actions and scatters the states and rewards.\\
    
    \midrule
    
    \textbf{[DP-F] -- Central} \newline \newline
    replicate: fused \emph{actor/learner}, \emph{env} \newline split: \emph{param server/policy pool}\label{fig:distributionParSever} &
    \includegraphics[align=t, width=\linewidth]{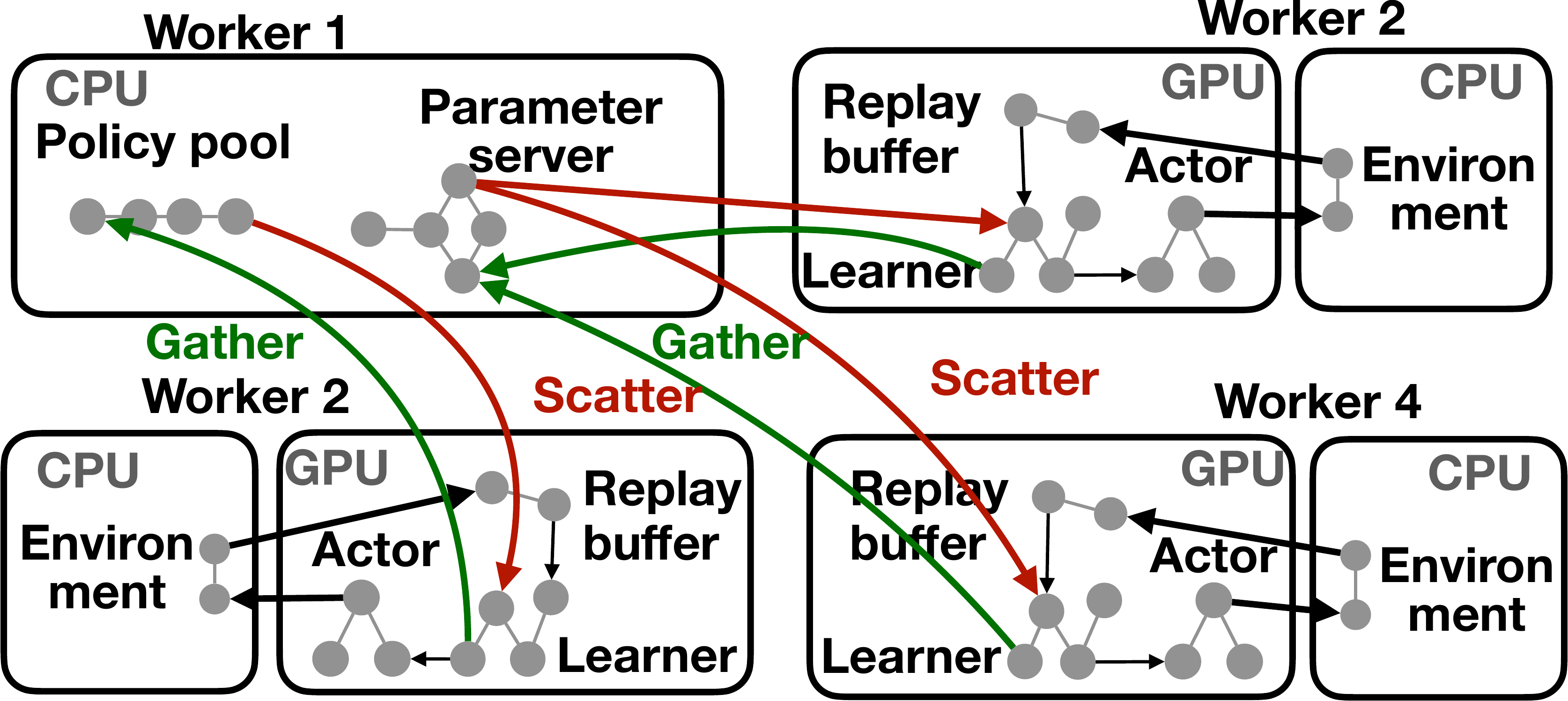} & DP-F supports a central \emph{policy pool}~\cite{DBLP:journals/corr/abs-2106-07551} or \emph{parameter server}~\cite{DBLP:conf/osdi/LiAPSAJLSS14} on a separate worker (W1). W2--W4 co-locate GPU fragments for policy inference and training and CPU fragments for environment execution.\\
    
    \bottomrule
  \end{tabular}
  \caption{Default distribution policies supported by \sys}\label{tab:distributionPolicies}
\end{table*}


\subsection{Generating and dispatching fragments}

The coordinator has two components:  

\myparr{\textnormal{The} FDG Generator} uses the FDG generation algorithm~(\A\ref{alg:build_fdg}) to partition the RL algorithm based on the partition annotations and to insert code for the interfaces.

To ensure the code is executable by a given DL engine, a user must implement parts of the RL algorithm using appropriate dataflow operators, such as \code{Mul()}, \code{Div()} etc. For the generated fragment interfaces, \sys maintains a mapping from the operations defined by the partition annotations to the corresponding dataflow operators. For example, the \code{AllGather} annotation in \A\ref{alg:mappo} maps to a \code{comms.AllGather} operator~\cite{comms_allgather} in the MindSpore DL engine.

The generated code is synthesised as part of the \code{run()} method of a \code{Fragment} template class. A worker then executes the fragment by invoking this method.

\myparr{\textnormal{The} Fragment Dispatcher} launches multiple instances of the DL engine on each worker, and sets up the distributed infrastructure for communication, \eg through MPI~\cite{mpi40}. It also assigns fragments to devices according to the specified distribution policy and sends the fragments to the workers.

\subsection{Executing fragments on workers}

Each worker also has two components:

\myparr{\textnormal{The} Code Generator} produces executable code that is specific to a given execution environment, such as DL engine, for the fragments received from the coordinator. Based on the fragment's AST, it also synthesises the interface implementations.
Finally, it converts the AST to Python code using the \code{unparse()} function from the Python AST library.

When a worker is assigned multiple fragment replicas on a single GPU, it uses several CUDA streams concurrently, one for each fragment. This approach, however, incurs an overhead launching and scheduling streams and also leads to extra memory copies between the CPU and GPU. \sys avoids this by \emph{fusing} co-located fragments into a single one. This leverages the support of DL engines to process data in a SIMD fashion: by batching tensors from multiple replicated fragments into a single tensor, they can be processed by the DL engine's data-parallel operators.

To fuse fragments, the Code Generator locates the AST node for the tensor in the fragment. Since the Python AST represents a tensor's shape and data as a list stored in the \code{elts} field, the Code Generator can create the appropriate tensor shape for a batched operator.

\myparr{\textnormal{The} Executor} launches the generated fragment code on different devices based on their implementations. For example, code implemented using MindSpore operators can run on a GPU controlled by the MindSpore DL engine; pure Python fragments execute on CPU cores. For efficient GPU execution, DL engines such as MindSpore compile code into a computational graph with GPU kernels. This enables optimisations: \eg independent operators can be scheduled in parallel, improving performance over sequential execution. 

Communication between fragments is also handled by MindSpore operators, which automatically select a suitable implementation based on the device type and network topology. It may use NCCL~\cite{nccl} for multi-GPU collective communication and MPI~\cite{mpi40} for InfiniBand traffic between workers.



\section{Distribution Policies}
\label{sec:distribution}

A distribution policy governs how \sys executes an RL algorithm by specifying the FDG partitioning and the mapping of fragments to workers and devices. In general, there is not a single policy that is optimal in all cases: the performance and applicability of a policy depends on the type of RL algorithm, the size and complexity of the DNN model, its hyper-parameters, the available cluster compute resources (\ie CPUs and GPUs), and the network bandwidth. \sys allows users to switch easily between policies, either for the same RL algorithm or when using different algorithms.

\T\ref{tab:distributionPolicies} describes \sys{}'s six~default policies, showing the trade-off between them. These policies largely follow the hard-coded distribution strategies of existing RL systems; further policies can be defined easily by expert users:

\tinyskip

\noindent
\emph{DP-A (Single learner/coarse)} replicates the actor and environment fragments but relies on a single learner. The DNN policy is replicated across the actors and learner, which only requires coarse synchronisation. This policy is therefore suitable for expensive environments that need scaling out, but small DNN models that can be synchronised in a batched fashion.

\tinyskip

\noindent
In contrast, \emph{DP-B (Single learner/fine)} fuses the actor and environment into a single CPU fragment, and only deploys the learner on a GPU. Therefore it does not communicate policy parameters between workers, which is preferable for large DNN models with many parameters. Compared to the DP-A, it relies on fine-grained synchronisation: training data is exchanged at each step, instead of being batched per episode. For good performance, DP-B therefore requires high bandwidth connectivity between workers.

\tinyskip

\noindent
\emph{DP-C (Multiple learners)} performs data-parallel training with multiple learners. This policy is necessary when the data generated from actors becomes too large for a single GPU, and \eg DP-A cannot be used. However, it requires the tuning of hyper-parameters (\eg the learning rate) to scale due to its reliance on data parallelism. Since workers only exchange information about the trained policy (\eg aggregated DNN gradients), DP-C is communication efficient.

\tinyskip

\noindent
\emph{DP-D (GPU only)} fuses the full RL training loop into a single GPU fragment. It is only applicable if the environment has a GPU implementation. Since the full RL training loop is accelerated, DP-D achieves the highest performance by eliminating the overhead of CPU/GPU switches.

\tinyskip

\noindent
\emph{DP-E (Environments)} dedicated one or more workers for the execution of environments. It is best suited when the RL training job executes computationally expensive CPU environments, \eg complex physics simulations, which requires many dedicated CPU cores for execution.

\tinyskip

\noindent
Finally, \emph{DP-F (Central)} introduces a separate fragment for a centralised policy pool~\cite{DBLP:journals/corr/abs-2106-07551} or parameter server~\cite{DBLP:conf/osdi/LiAPSAJLSS14}. It can support CTDE-based MARL algorithms (similar to Ray~\cite{DBLP:conf/osdi/MoritzNWTLLEYPJ18} and RLlib~\cite{DBLP:conf/icml/LiangLNMFGGJS18}) by workers sending local policy updates to a parameter server; or population-based MARL (similar to MALib~\cite{DBLP:journals/corr/abs-2106-07551}) by a worker maintaining a population of policies, selecting the optimal one and broadcasting it to other workers.


\renewcommand*{\thefootnote}{\arabic{footnote}}
\section{Evaluation}
\label{sec:evaluation}

\newcommand{\dpone}{DP-B\xspace}
\newcommand{\dptwo}{DP-A\xspace}
\newcommand{\dpthree}{DP-C\xspace}
\newcommand{\dpfour}{DP-D\xspace}
\newcommand{\dpfive}{DP-E\xspace}
\newcommand{\dponeL}{DP-B'\xspace}
\newcommand{\dptwoL}{DP-A'\xspace}

Our experimental evaluation answers the following questions: ``what are the trade-offs between distribution policies?''~(\S\ref{sec:eval:trade_offs}); ``what is \sys{}'s performance compared to other systems?''~(\S\ref{sec:eval:overhead}); and ``how well does \sys scale?''~(\S\ref{sec:eval:scalability}).

\subsection{Experimental set-up}
\label{sec:eval:setup}

\mypar{Implementation and test-bed} We implement MSRL on top of MindSpore~\cite{mindspore}. The code is available at \url{https://github.com/mindspore-ai/reinforcement}. We conduct the experiments on both a \emph{cloud} and a \emph{local} cluster. As shown in \T\ref{tab:ClusterSetting}, the cloud cluster has 16~VMs; the local cluster has 4~nodes. All machines run Ubuntu Linux~20.04 with CUDA~11.03, cuDNN~8.2.1, OpenMPI~4.0.3 and MindSpore~1.8.0.

\mypar{RL algorithms and environments} We answer the above questions by focusing on three widely-used RL algorithms, Proximal Policy Optimization~(PPO)~\cite{DBLP:journals/corr/SchulmanWDRK17}, its multi-agent version, Multi-Agent PPO~(MAPPO)~\cite{mappo2021}, and Asynchronous Advantage Actor Critic~(A3C)~\cite{a3c2016}. The policy uses a seven-layer DNN. As environments, we use three games from MuJoCo~\cite{todorov2012mujoco}, an advanced physics simulation engine, and the Multi-Agent Particle Environment~(MPE)~\cite{DBLP:conf/nips/LoweWTHAM17}, a set of mixed cooperative-competitive environments for MARL.

\mypar{Distribution policies} With the above algorithms, we use the following distribution policies from \T\ref{tab:distributionPolicies}: \dptwo (Single learner/coarse), \dpone (Single learner/fine), \dpthree (Multiple learners), \dpfour (GPU only) and \dpfive (Environments).

\begin{table}[t]
  \centering
  \resizebox{\columnwidth}{!}{ 
    \begin{tabular}
      {llll}
      \toprule
      \textbf{Cluster} & \textbf{CPU cores} & \textbf{GPUs} & \textbf{Interconnects}\\ 
      &\emph{\#nodes} $\times$ \emph{\#per node} & \emph{\#nodes} $\times$ \emph{\#per node} & \emph{intra-, inter-node} \\
      \midrule
      \scell[c]{Azure VMs\\NC24s\_v2} & \scell[c]{Intel Xeon E5-2690\\16$\times$24, 448\unit{GB}} & P100, 16$\times$4 & PCIe, 10~GbE\\
      \midrule
      Local cluster & \scell[c]{Intel Xeon 8160\\ 4$\times$96,
      250\unit{GB}} & V100, 4$\times$8 & \scell[c]{NVLink,\\ 100\unit{Gbps} IB}\\
      \bottomrule
    \end{tabular}
  }
  \caption{Testbed configuration}\label{tab:ClusterSetting}
\end{table}

\mypar{Metrics} For PPO, we measure (i)~the training time to reach a given reward value and (ii)~the time for each episode. For MAPPO, as the problem size increases with the number of agents due to agent competition, we report training time against the problem complexity and the training throughput.

\subsection{Trade-offs between distribution policies}
\label{sec:eval:trade_offs}

\label{sec:eval:scaleEnv}

First we observe the impact of changes to the RL workload and hardware resources on different distribution policies.

\begin{figure}[tb]
  \centering
  \begin{subfigure}[t]{0.48\columnwidth}
    \includegraphics[width=\textwidth]{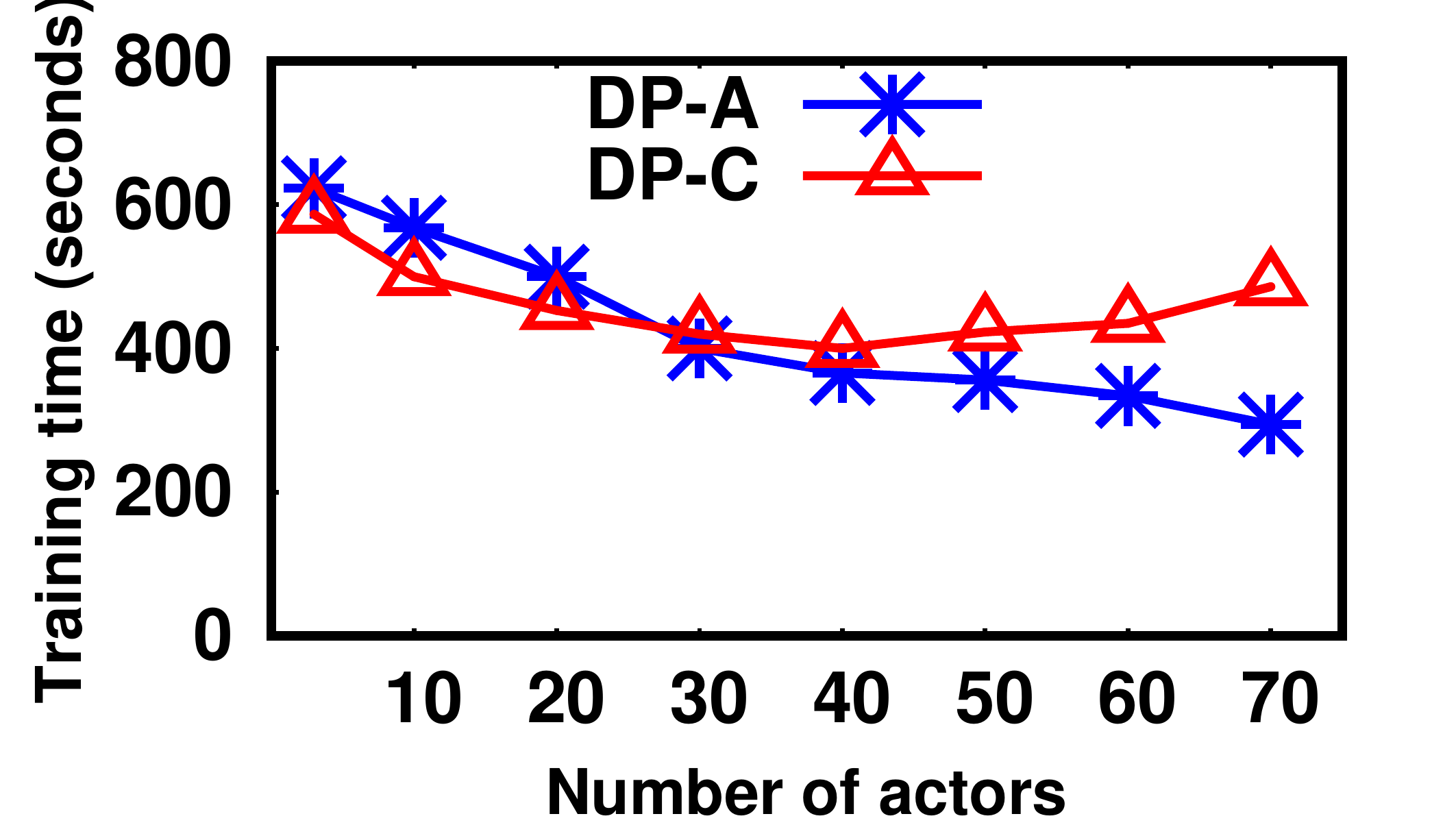}
    \caption{Training time vs. actors (PPO)}\label{fig:DP_actors}
  \end{subfigure}
  \begin{subfigure}[t]{0.48\columnwidth}
    \centering
    \includegraphics[width=\textwidth]{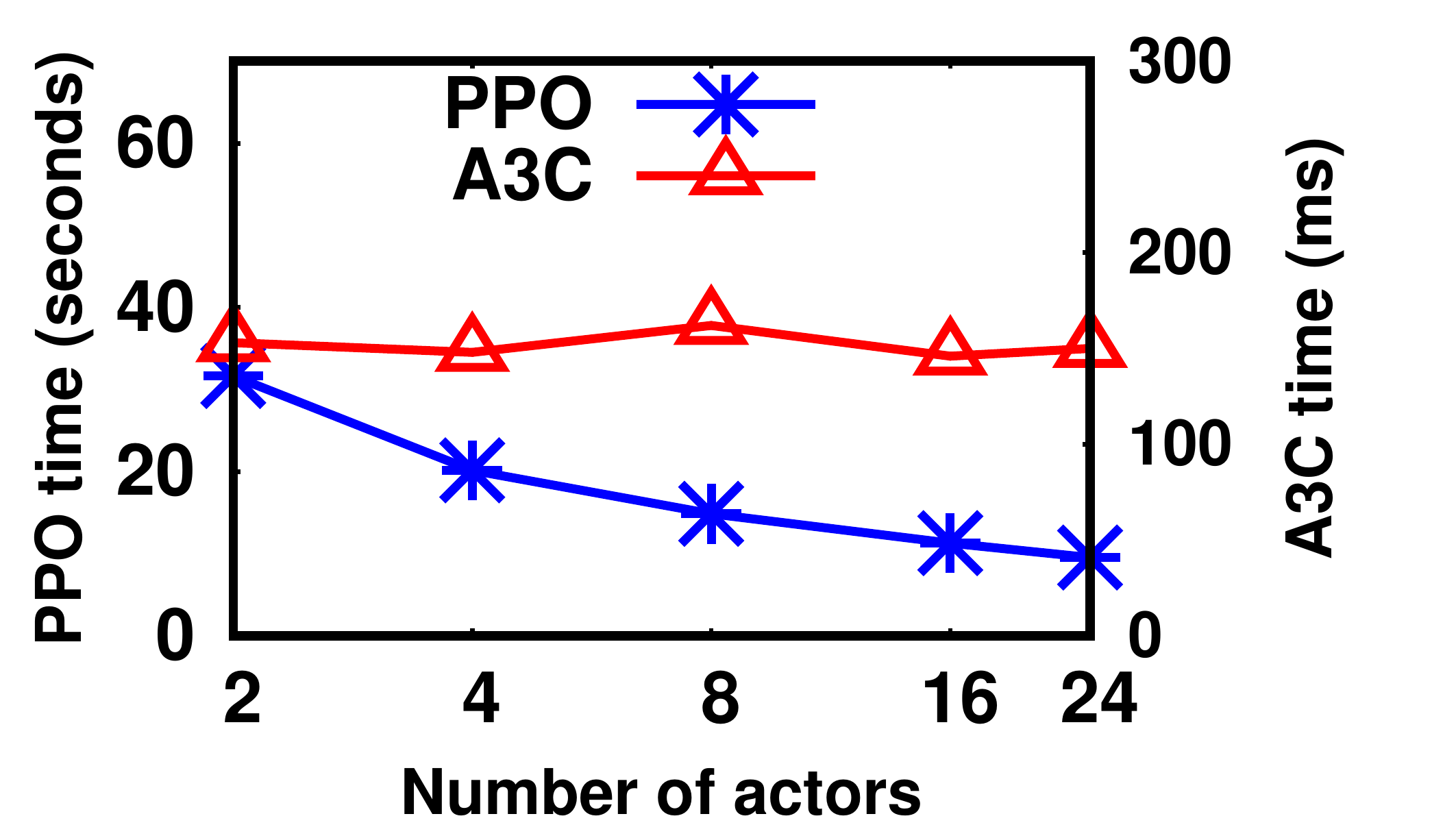}
    \caption{Episode time (PPO vs. A3C)}\label{fig:DP_ppo_a3c}
  \end{subfigure}
  \\
  \begin{subfigure}[t]{0.48\columnwidth}
    \includegraphics[width=\textwidth]{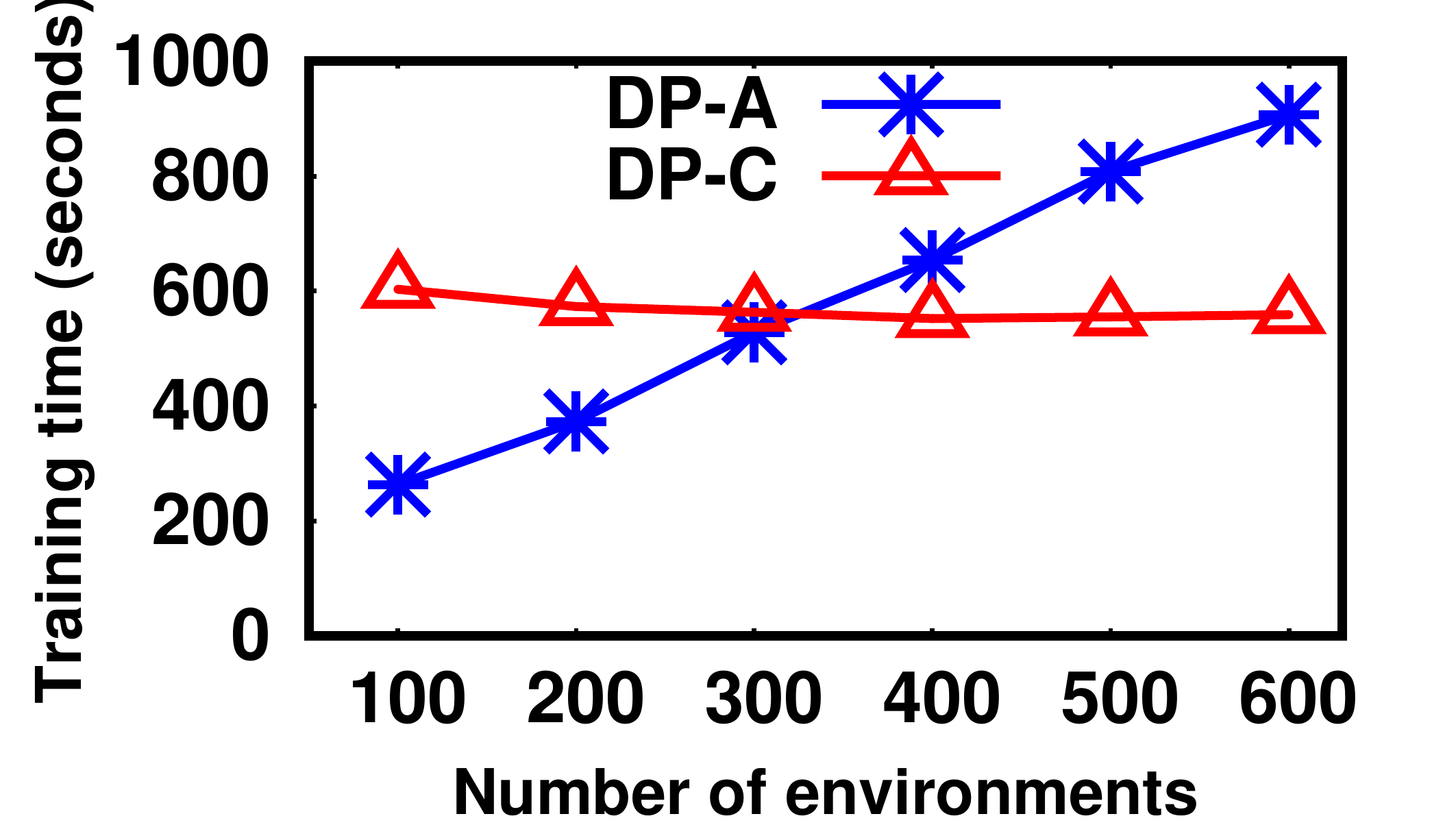}
    \caption{Training time vs. envs}\label{fig:DP_envs}
  \end{subfigure}
  \begin{subfigure}[t]{0.48\columnwidth}
    \centering
    \includegraphics[width=\textwidth]{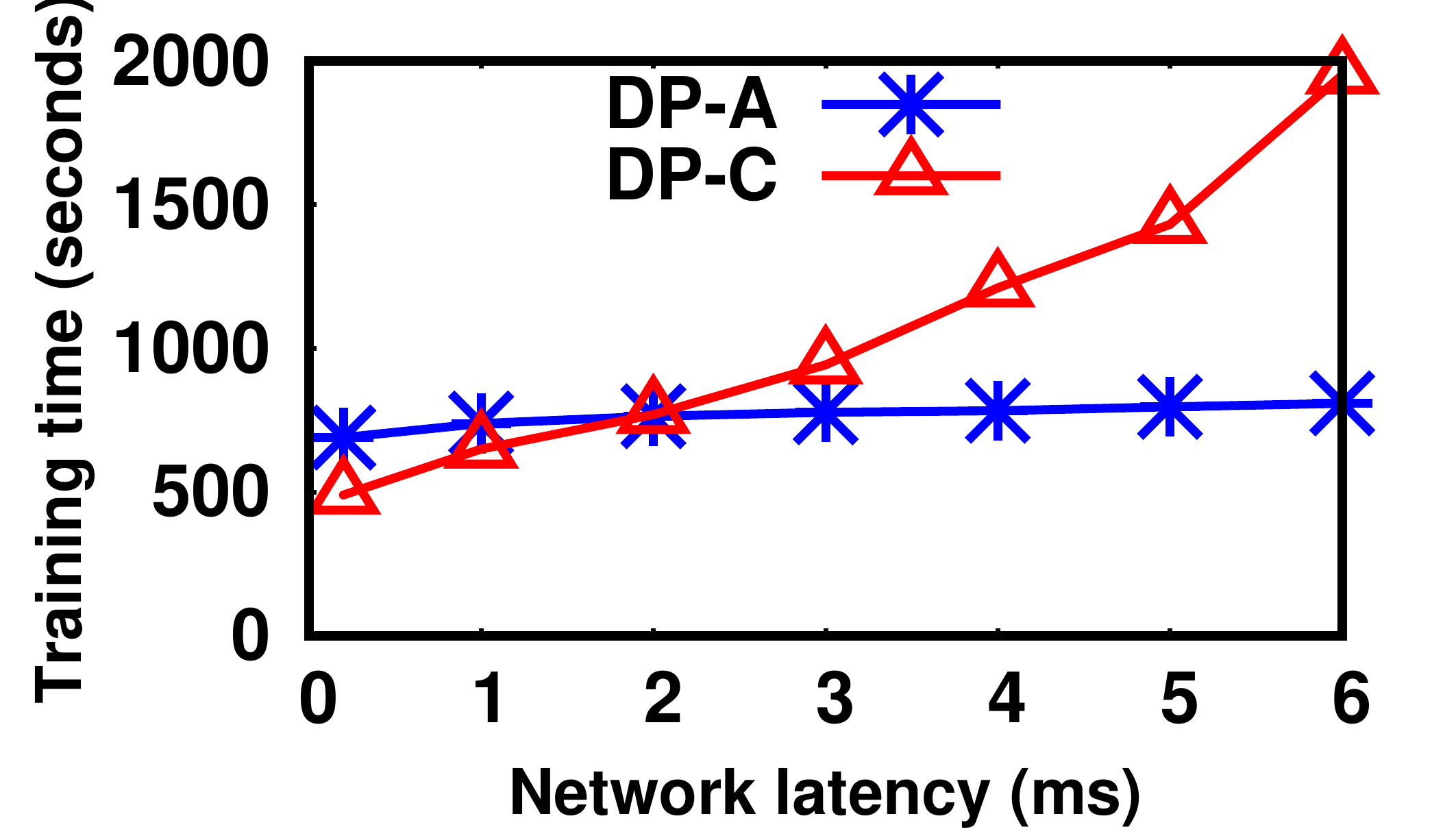}
    \caption{Training time vs. network latency}\label{fig:DP_lat}
  \end{subfigure}
  \caption{Impact of hyper-parameters and network properties}\label{fig:DPs}
\end{figure}

\mypar{Actors} We investigate the impact of the number of actors used when training PPO to a reward of 3,000 with 200~environments. We report the trade-off between two policies, \dptwo (Single learner/coarse) and \dpthree (Multiple learners). 

\F\ref{fig:DP_actors} shows the training time with 2 to 70~actors. \dptwo scales better than \dpthree with more actors: it converges faster with a higher actor count. \dpthree achieves the best performance with 40~actors, and its performance deteriorates after that but it outperforms \dptwo with fewer than 30~actors.

Since \dptwo only has 1~learner, its training batch size is fixed. Increasing the number of actors therefore only distributes environment execution. In contrast, \dpthree fuses actors and learners into single fragments. With more actors, it also adds learners and thus reduces the batch size for each learner. This adds randomness to the training~\cite{DBLP:conf/nips/HofferHS17} and affects converge speed. In general, \dpthree needs further hyper-parameter tuning (\eg of the learning rate) for better performance~\cite{DBLP:conf/nips/HofferHS17}.

Next we compare the behaviour of two algorithms, PPO and A3C, under the same distribution policy~\dptwo. \F\ref{fig:DP_ppo_a3c} shows the time per episode for up to 24~actors. (Note that A3C requires at least 2 actors.) Under \dptwo, PPO's episode time decreases as the number of actors increases; in contrast, A3C's episode time stays constant with the actor count.

In PPO, adding actors increases the parallelism degree of environment execution and therefore reduces the workload per actor; in A3C, each actor only interacts with one environment, making its workload independent of the actor count. To reduce the time per episode for A3C, a new distribution policy could be written that distributes the actor among multiple devices, \eg exploiting data- or task-parallelism.

\mypar{Environments} We investigate the impact of changing the number of environments. When an agent interacts with more environments in parallel within one episode, it trains with more data, thus potentially improving training performance. We fix the number of actors to 50.

\F\ref{fig:DP_envs} shows the training time as we vary the number of environments from 100 to 600 under \dptwo and \dpthree. \dpthree scales better than \dptwo with more environments, and there is a cross-over point with 320~environments. The training time with \dptwo increases with more environments but remains stable for \dpthree. This is because \dptwo's actors send trajectories to the learner, which increases in communication overhead with more environments. In contrast, \dpthree only communicates gradients and its communication overhead is fixed. Therefore the right distribution policy must be selected based on the environment count.

\mypar{Network latency} We also examine the behaviour of \dptwo and \dpthree when deployed with PPO on clusters with different network latencies. We change the network latency in our cloud cluster using the Linux traffic control (\texttt{tc}) tool from 0.2\unit{ms} to 6\unit{ms}. We use 400~environments and 50~actors.

\F\ref{fig:DP_lat} reports training time. \dpthree is more sensitive to network latency than \dptwo. Its training time increases rapidly with higher network latency, while remaining stable for \dptwo. Since \dpthree leverages Mindspore's data parallel model~\cite{mindspore} to broadcast, aggregate and update gradients, it repeatedly transmits many small tensors; \dptwo transmits the trajectory and DNN model weights as compact large tensors, performing data transmissions less frequently. However, \dpthree transmits less data than \dptwo and thus is more appropriate in clusters with low network latencies (below 2\unit{ms}).

\begin{figure}[tb]
  \begin{subfigure}[t]{0.49\columnwidth}
    \centering
    \includegraphics[width=\textwidth]{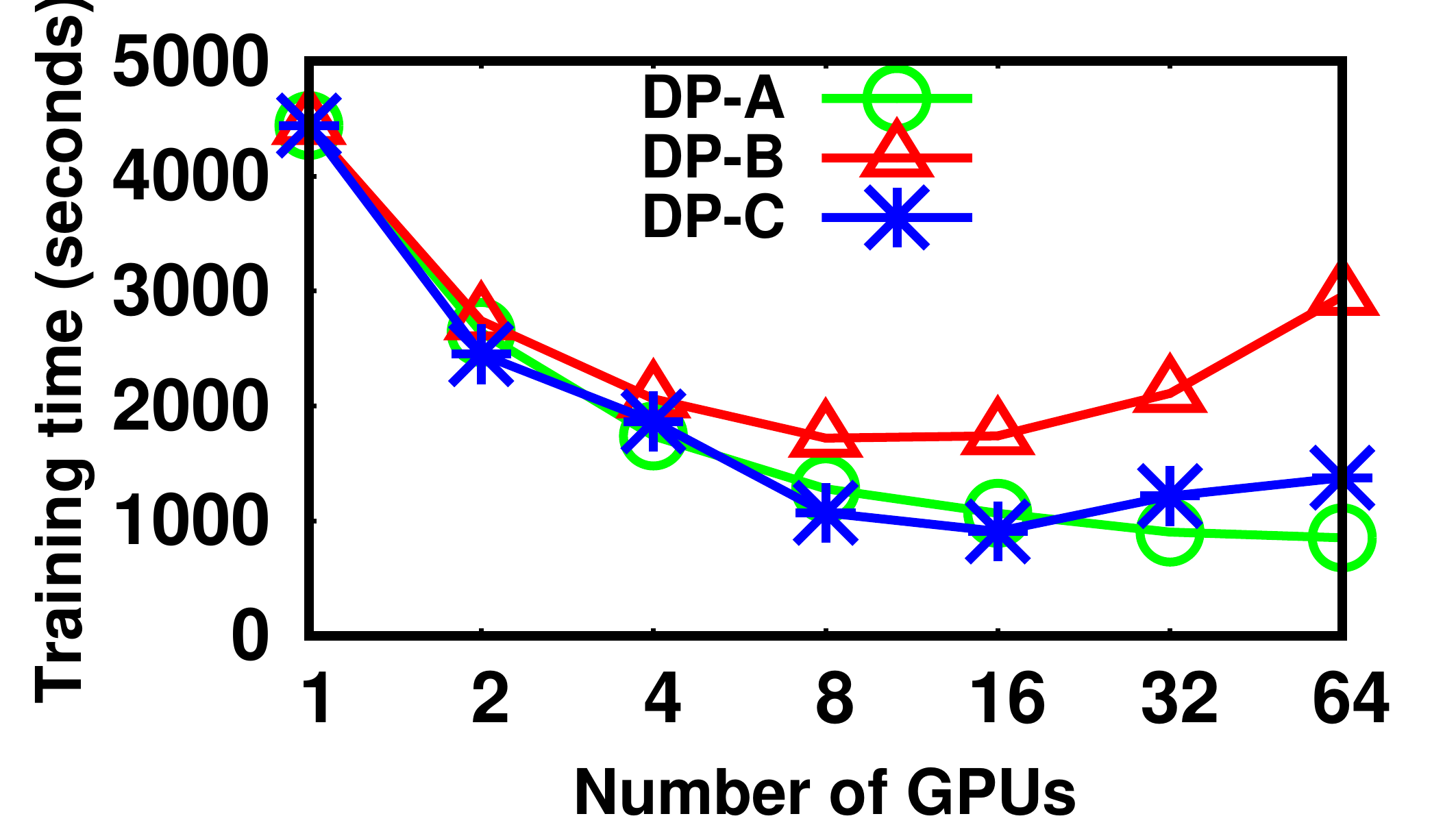}
    \caption{Training time vs. GPUs (cloud)}\label{fig:ScalePPO}
  \end{subfigure}
  \begin{subfigure}[t]{0.49\columnwidth}
    \centering
    \includegraphics[width=\textwidth]{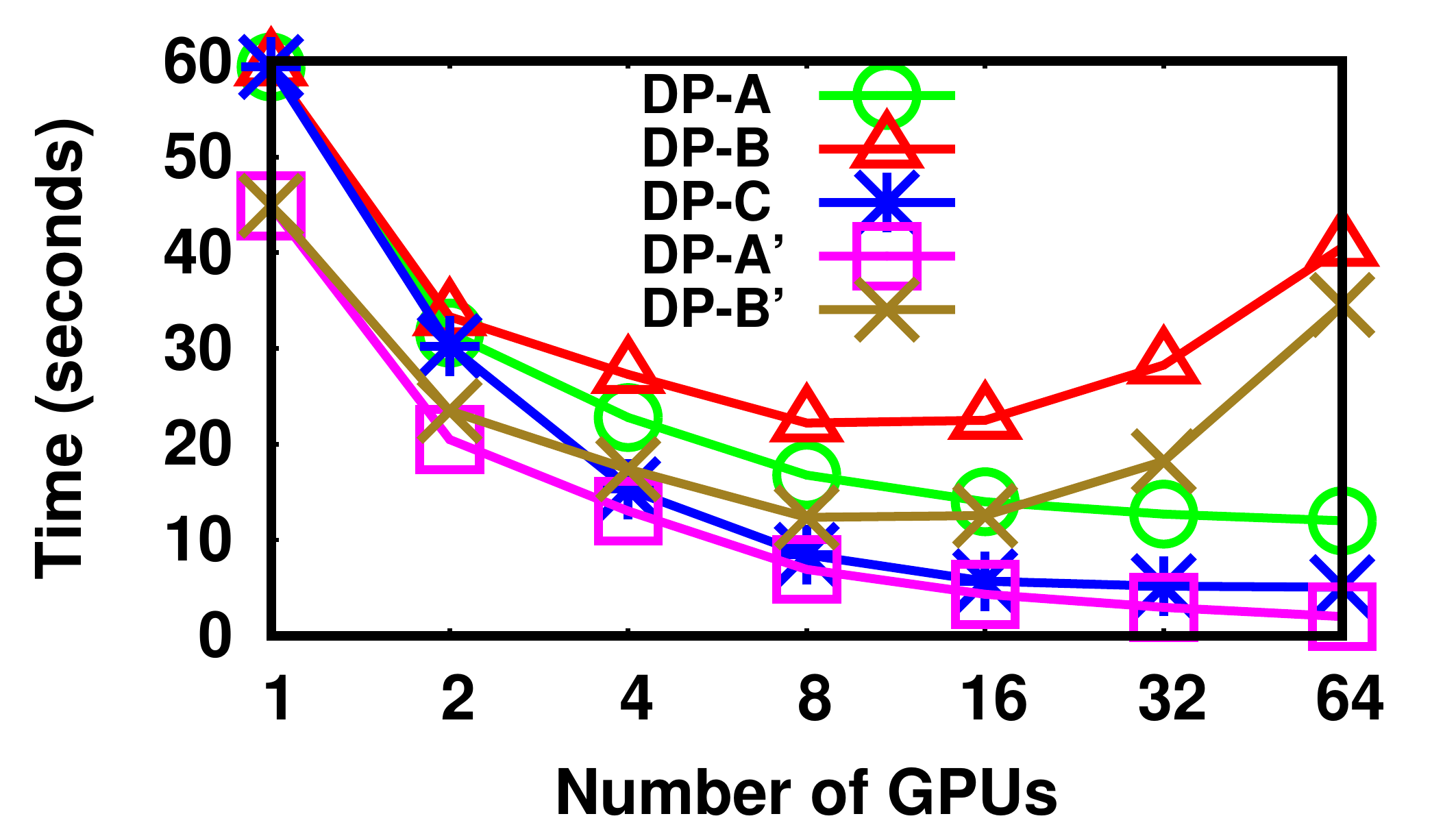}
    \caption{Episode time vs. GPUs (cloud)}\label{fig:ScalePPOEpoch}
  \end{subfigure}
  \\
  \begin{subfigure}[t]{0.49\columnwidth}
    \centering
    \includegraphics[width=\textwidth]{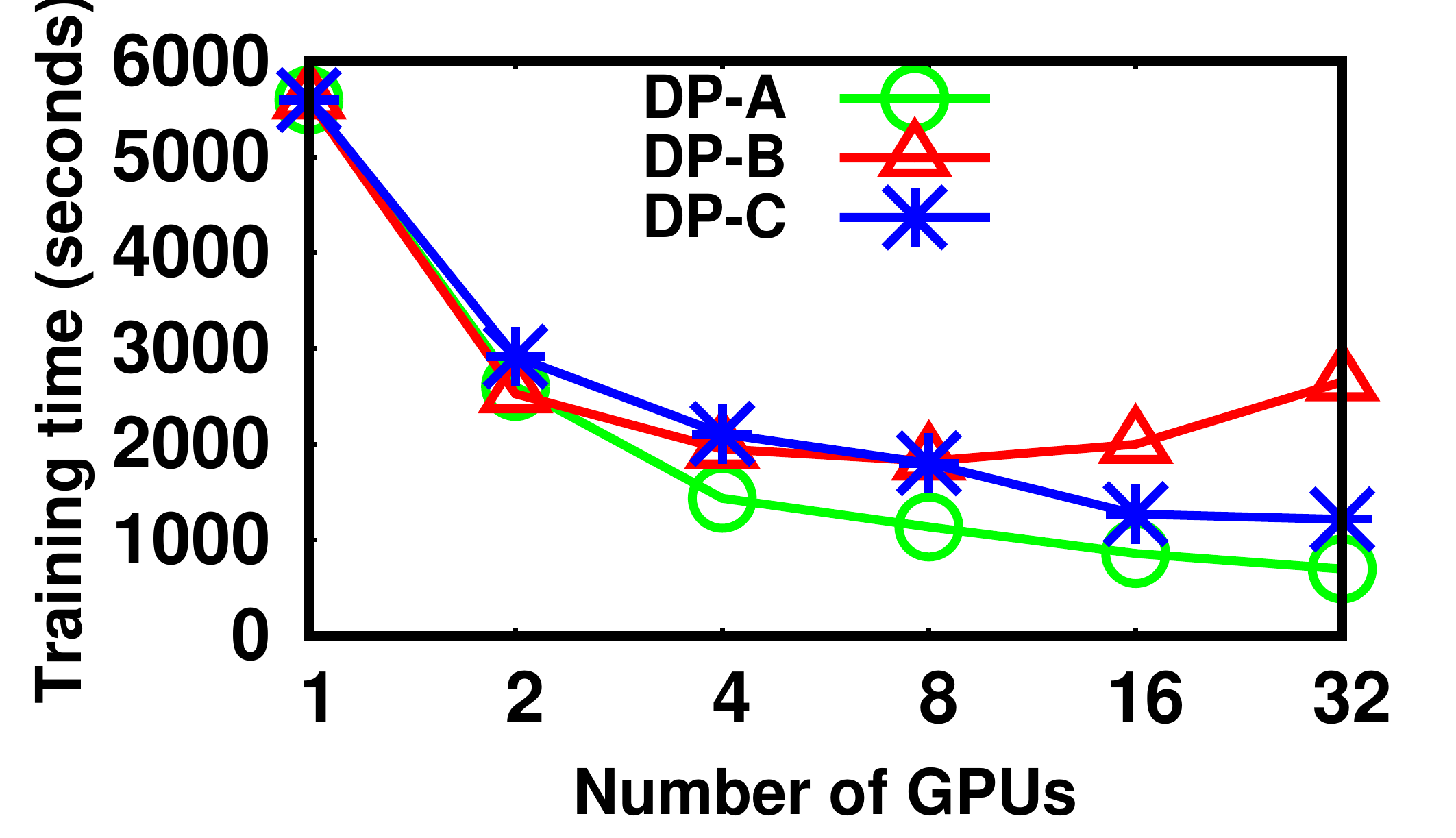}
    \caption{Training time vs. GPUs (local)}\label{fig:ScalePPOLocal}
  \end{subfigure}
  \begin{subfigure}[t]{0.49\columnwidth}
    \centering
    \includegraphics[width=\textwidth]{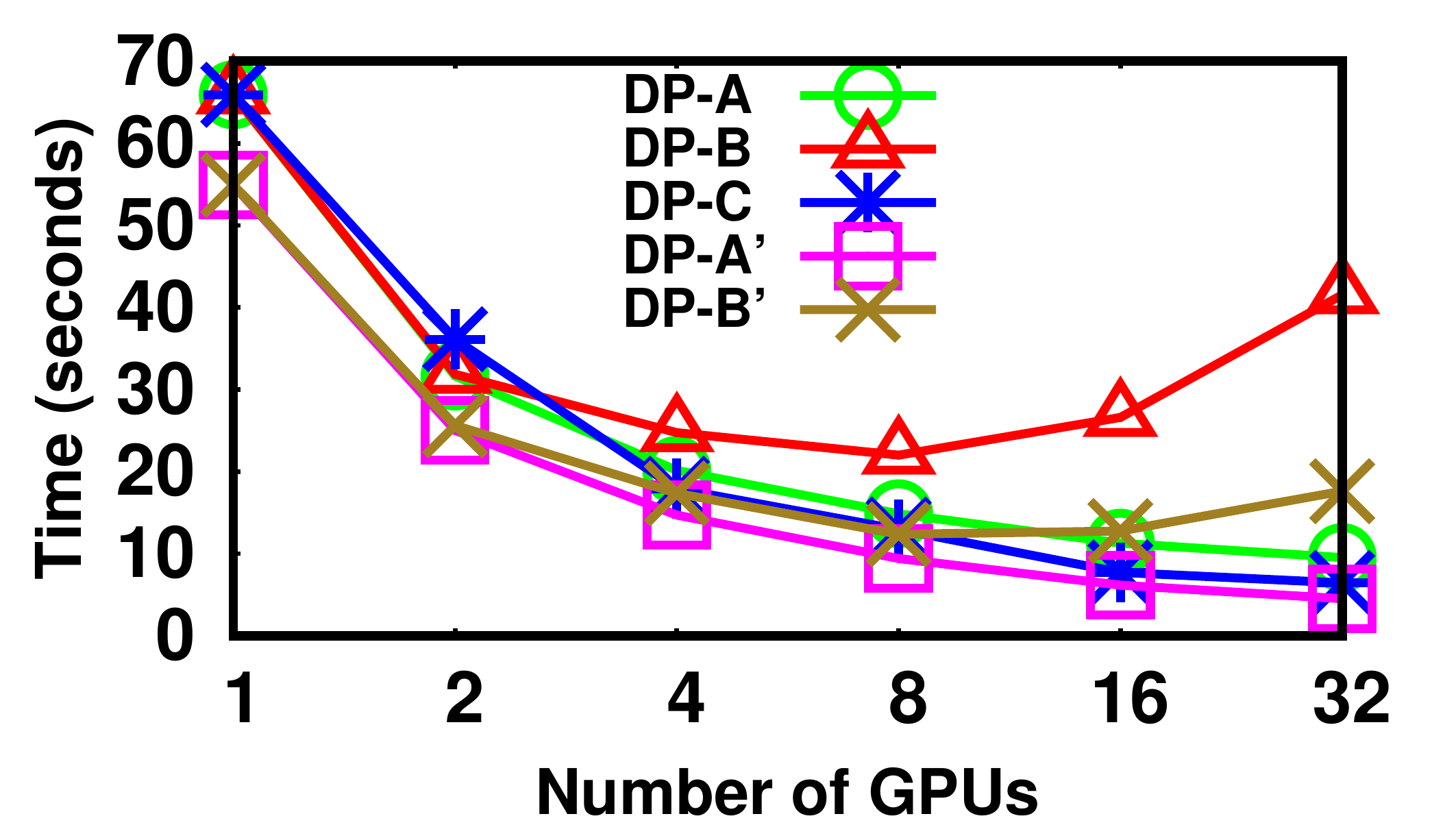}
    \caption{Episode time vs. GPUs (local)}\label{fig:ScalePPOEpochLocal}
  \end{subfigure}
  \caption{Impact of GPU number on training time}\label{fig:PPO_GPU_count}
\end{figure}

\mypar{Cluster size} Finally, we compare the performance of PPO under 3~distribution policies when increasing the number of GPUs (\F\ref{fig:PPO_GPU_count}): \dptwo and \dpone use a single learner but apply different synchronisation granularities; \dpthree scales to multiple learners using data-parallelism. We use a fixed number of 320~Mujoco HalfCheetah~\cite{gym} environments.

\F\ref{fig:ScalePPO} shows the training time to reach a reward of 4,000 in the cloud cluster with up to 64~GPUs; \F\ref{fig:ScalePPOEpoch} reports the time per episode. Compared to 1~GPU, all distribution policies reduce training time with more GPUs. With 64~GPUs, \dptwo achieves the highest speed-up in training time (5.3$\times$). Compared to that, \dpone exhibits a longer training time, which is lowest for 16~GPUs. \dptwo maintains local copies of the DNN model at the actor and learner, and only actors send the batched states and rewards to the learner at the end of each episode (\ie after 1,000~steps). This reduces the overhead with a large number of GPUs compared to \dpone, whose actor fragments must communicate with the learner at each step.

\dpthree exhibits a different behaviour: with 16~GPUs, it achieves better performance than either \dptwo and \dpone. This is because it distributes policy training: it trains smaller batches of trajectories on each device and aggregates the gradients from all devices to update the policy. Instead, \dpone and \dptwo gather all batches and train them using 1~learner.

While \dpthree is the best policy choice for 16~GPUs, it performs worse than \dptwo for a larger cluster. With more GPUs, trajectory batches become smaller and aggregating the trained gradients becomes less efficient compared to training a single large batch. Therefore, although \dpthree trains each episode faster than \dptwo (see~\F\ref{fig:ScalePPOEpoch}), it requires more episodes to reach a similar reward value.

Note that \dptwo and \dpone use the original implementation of the PPO algorithm~\cite{DBLP:journals/corr/SchulmanWDRK17}, which uses a single learner to train the policy. As a result, scalability is constrained by the centralised policy training (\myc{3} in \F\ref{fig:MARLTrain}). To ignore this bottleneck in the algorithm, we also exclude the policy training time in \F\ref{fig:ScalePPOEpoch}, labelled \dptwoL and \dponeL. Here, \sys continues scaling even with large GPU numbers: when moving from 32 to 64~GPUs, performance increases by 25\%.

In \Fs\ref{fig:ScalePPOLocal} and \ref{fig:ScalePPOEpochLocal}, we repeat the same experiment on the local cluster, which has faster GPU and network connectivity (NVLink, InfiniBand). When comparing \Fs\ref{fig:ScalePPOLocal} and~\ref{fig:ScalePPO}, we can see that \dptwo now always performs better than \dpthree, irrespective of the GPU count. Since the local cluster supports faster GPU communication, \dptwo's overhead is lower than in the cloud cluster. \dptwo{}'s larger batches become more effective than \dpthree's training/aggregation over smaller batches.

\myparii{Conclusions:} As the GPU count, hyper-parameters or network properties change, the differences between the synchronisation granularity and frequency of distribution polices impact performance. \sys{}'s ability to select a distribution policy at deployment time allows users to achieve the best performance in different configurations without having to change the algorithm implementation. 

\begin{figure}[tb]
  \centering
  \begin{subfigure}[t]{0.49\columnwidth}
    \includegraphics[width=\textwidth]{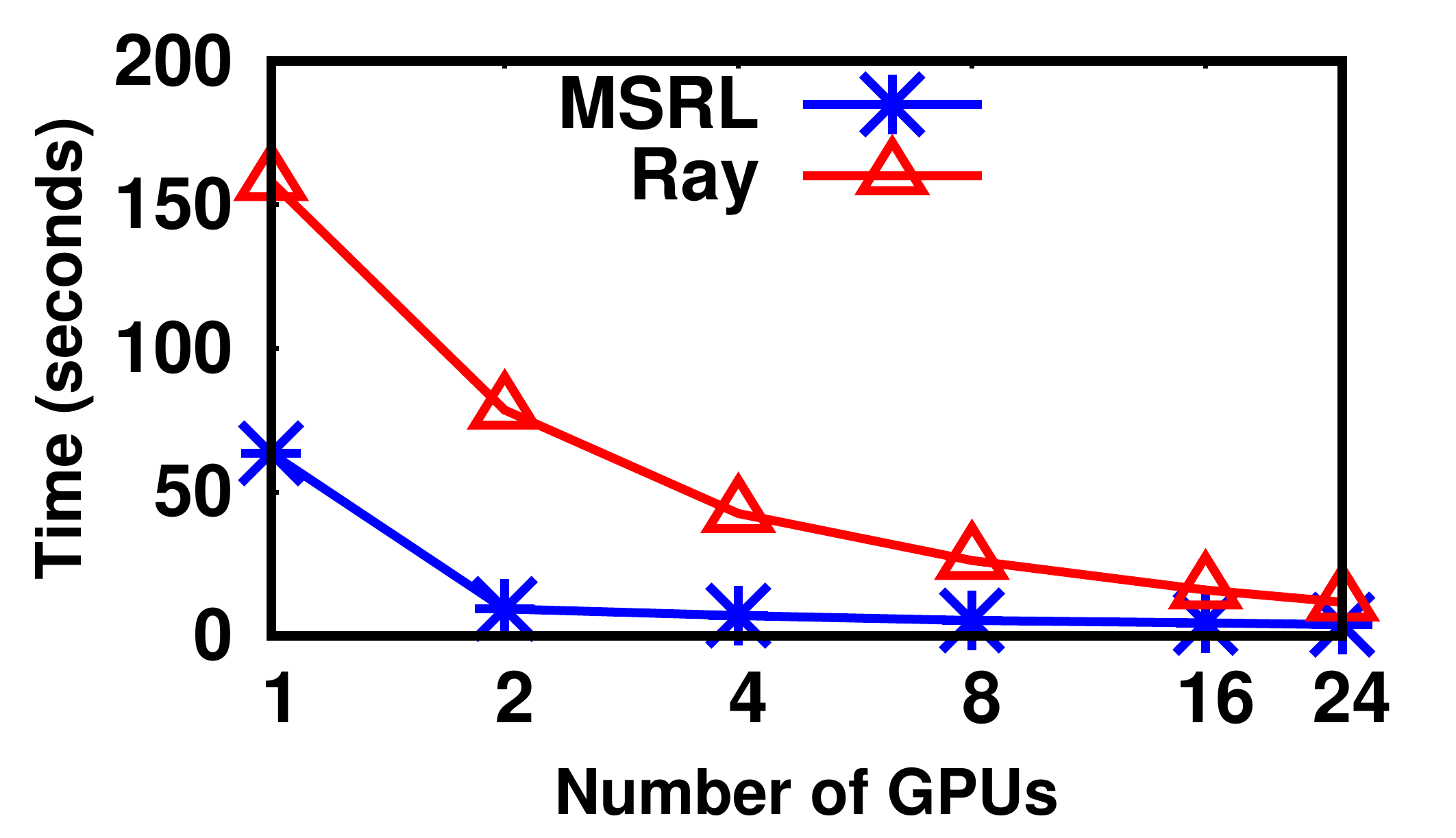}
    \caption{Episode time vs. GPUs (PPO)}\label{fig:ray_ppo}
  \end{subfigure}
  \begin{subfigure}[t]{0.49\columnwidth}
    \includegraphics[width=\textwidth]{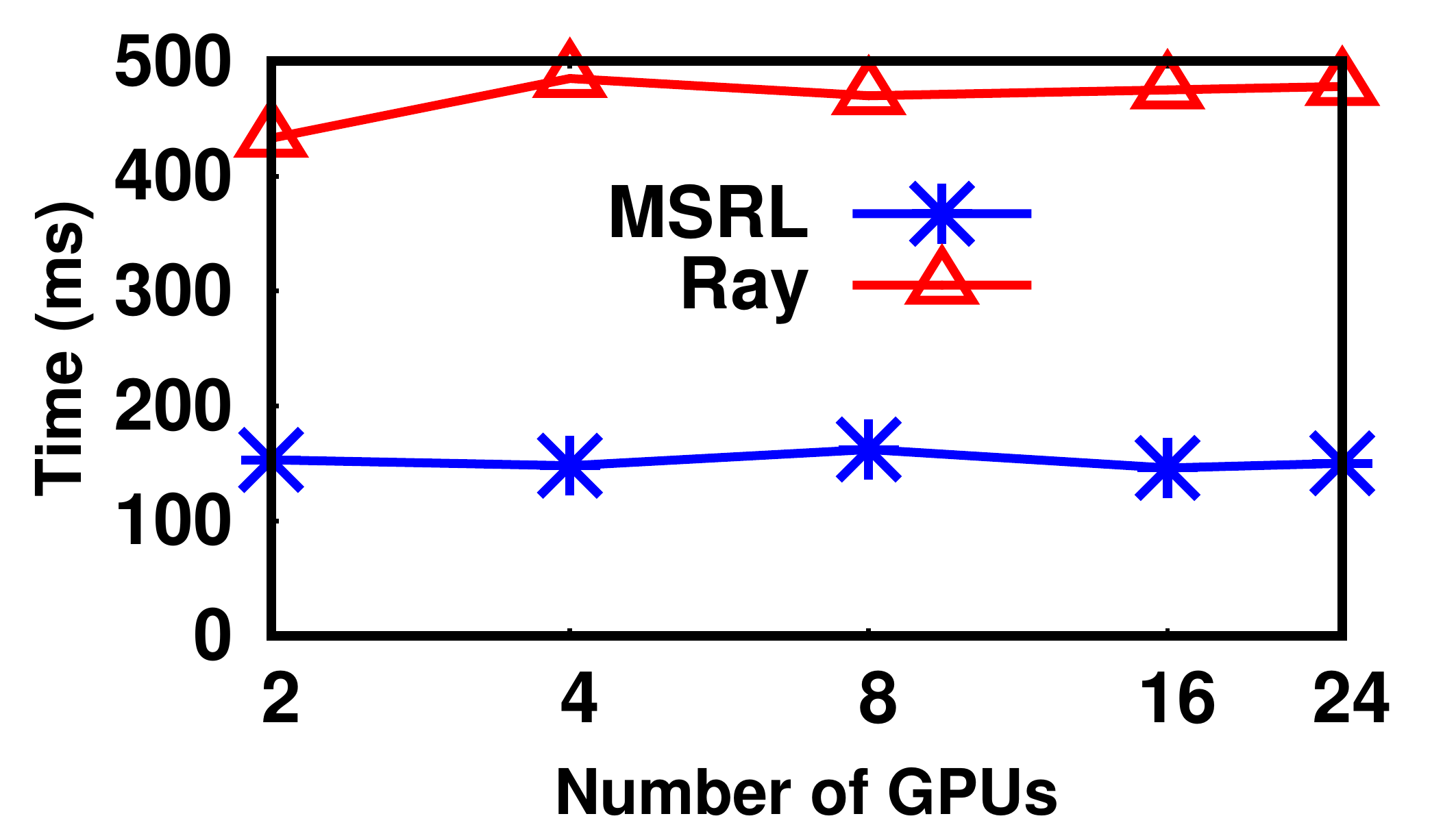}
    \caption{Episode time vs. GPUs (A3C)}\label{fig:ray_a3c}
  \end{subfigure}
  \caption{Performance comparison with Ray}\label{fig:rayEval}
\end{figure}

\subsection{Performance against baseline systems}
\label{sec:eval:overhead}

We want to put the absolute performance of \sys into perspective by comparing against other RL systems that hard-code particular parallelisation and distribution strategies. For the comparison, we choose Ray~\cite{DBLP:conf/osdi/MoritzNWTLLEYPJ18}, as a representative distributed RL training system, and WarpDrive~\cite{DBLP:journals/corr/abs-2108-13976}, as a single-GPU system that parallels the complete RL training loop.

\mypar{Distributed training} We compare against Ray~\cite{DBLP:conf/osdi/MoritzNWTLLEYPJ18} using PPO and A3C with DP-A on the local cluster. Both algorithms are implemented and tuned based on Ray{}'s public PyTorch-based implementation~\cite{raysetting}. We measure the time per episode, which is dominated by actor and environment execution. Since the time spent on DNN training/inference is negligible, the fact that \sys and Ray use different DL frameworks (MindSpore vs. PyTorch) has low impact. 

\F\ref{fig:ray_ppo} shows PPO's time per episode. In this experiment, we distribute 320~environments evenly among the actors, \ie each actor interacts with $320 / \textit{\#actors}$ environments, and a single learner trains the DNN. As shown in the figure, \sys{}'s episode time with 1~GPU is $2.5\times$ faster than Ray's because Ray executes the actor on the CPU, which then interacts with all environments sequentially. As the number of GPUs increases, both systems reduce episode time as each actor interacts with fewer environments. With 24~GPUs, it takes 3.85\unit{s} for \sys to execute an episode compared to 11.38\unit{s} for Ray ($3\times$ speed-up). When actors interact with multiple environments, \sys combines DNN inference into one operation through FDG fusion, exploiting GPU parallelism better. It also uses fragments to execute environment steps in parallel by launching multiple processes.

\F\ref{fig:ray_a3c} shows A3C's time per episode. Here, 1~learner performs gradient optimisation with gradients collected asynchronously from actors. Each actor interacts with 1~environment and computes gradients locally. The figure shows that both systems' time per episode remains constant with more GPUs. Each GPU is mapped to one actor, and its workload remains unchanged. Again, \sys executes actors $2.2\times$ faster than Ray: since its distribution policy exploits asynchronous \code{send}/\code{receive} operations from NCCL, \sys avoids further data copies between GPU and CPU. In contrast, Ray copies data to the CPU to communicate asynchronously.

\begin{figure}[tb]
  \centering
  \begin{subfigure}[t]{0.49\columnwidth}
    \includegraphics[width=\textwidth]{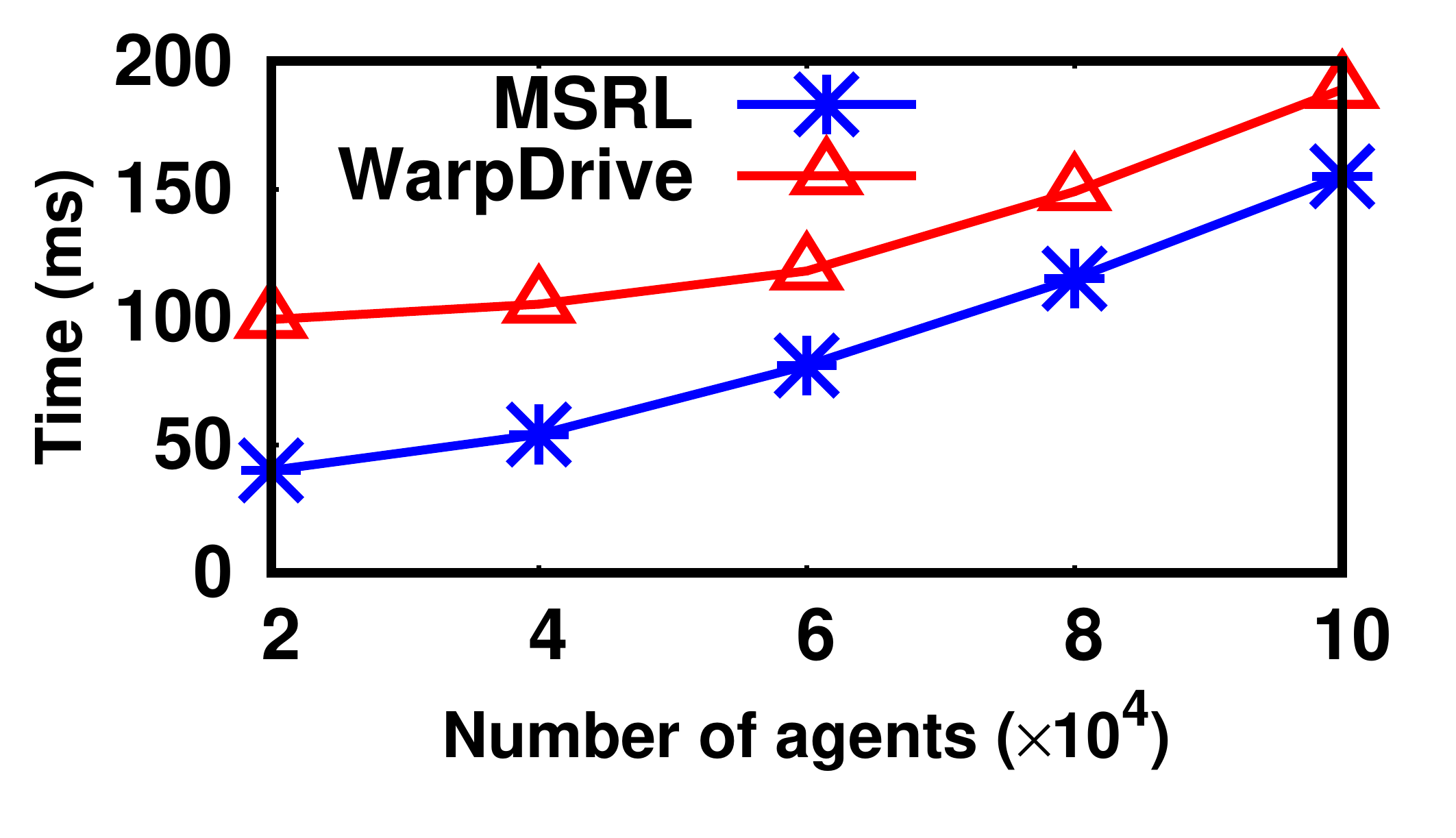}
    \caption{Episode time vs. agents (1~GPU)}
    \label{fig:compareTag}
  \end{subfigure}
  \begin{subfigure}[t]{0.49\columnwidth}
    \includegraphics[width=\textwidth]{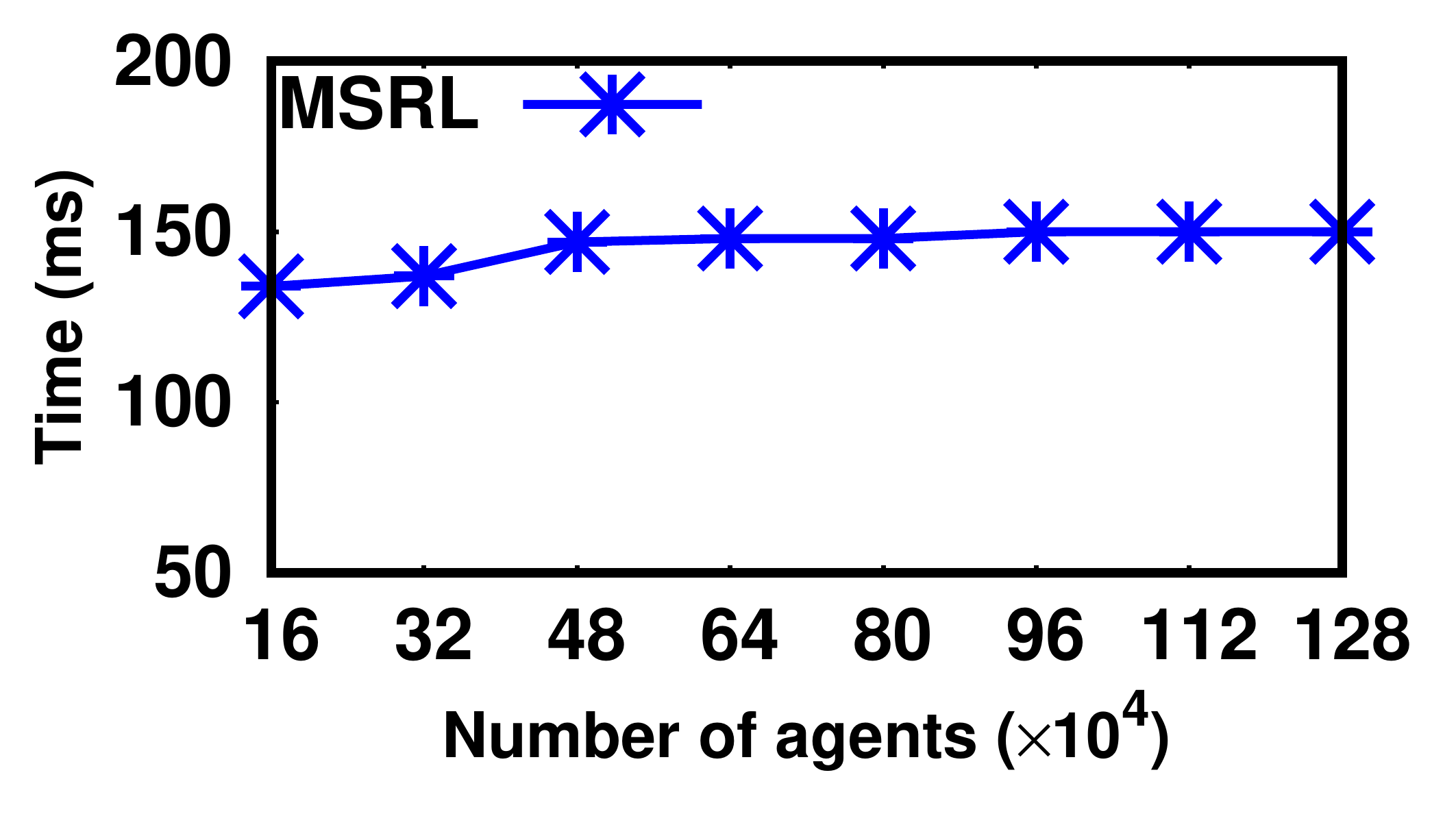}
    \caption{Episode time vs. agents (GPUs)}
    \label{fig:mgpuTag}
  \end{subfigure}
  \caption{Scalability of GPU-only PPO}\label{fig:ScalePPOTag}
\end{figure}

\mypar{GPU only training} Next, we deploy PPO with distribution policy \dpfour, which fuses the training loop into a single GPU fragment and replicates it for distributed execution. We use PPO with the MPE \emph{simple
  tag} environment~\cite{DBLP:conf/nips/LoweWTHAM17}, which simulates a predator-prey game in which chaser agents are rewarded for catching runner agents. Runner agents are penalised for being caught by a chaser or colliding with runners.

After adapting the \emph{simple tag} implementation to GPUs, we train different numbers of agents (thus increasing the number of environments) on the local cluster and measure the training time per episode. We compare against WarpDrive~\cite{DBLP:journals/corr/abs-2108-13976}, which performs single-GPU end-to-end RL training.

\F\ref{fig:compareTag} shows the training time on 1~GPU. Compared to WarpDrive, \sys is 1.2$\times$--2.5$\times$ faster when ranging from 20,000 to 100,000~agents. \sys{}'s DL engine (MindSpore) compiles fragments to computational graphs, exploiting more parallelisation and optimisation opportunities.

\F\ref{fig:mgpuTag} shows how \sys scales to multiple GPUs (which is unsupported by WarpDrive) when each GPU trains 80,000~agents. The training time increases from 138\unit{ms} to 150\unit{ms} from 160,000 to 960,000~agents. Beyond that, it remains stable -- the available connectivity bandwidth (NVLink, InfiniBand) now bounds the communication overheads between GPU fragments and devices.

\myparii{Conclusions:} \sys{}'s FDG abstraction provides the flexibility to use distribution policies for PPO and A3C that are tailored to their bottlenecks, \eg enabling parallel environment execution and aggressively parallelising GPU execution. Ray is limited by the distribution strategy of its RLlib library; WarpDrive's manual CUDA implementation prevents it from leveraging more sophisticated compiler optimisations.

\begin{figure}[tb]
  \centering
  \begin{subfigure}[t]{0.48\columnwidth}
    \includegraphics[width=\textwidth]{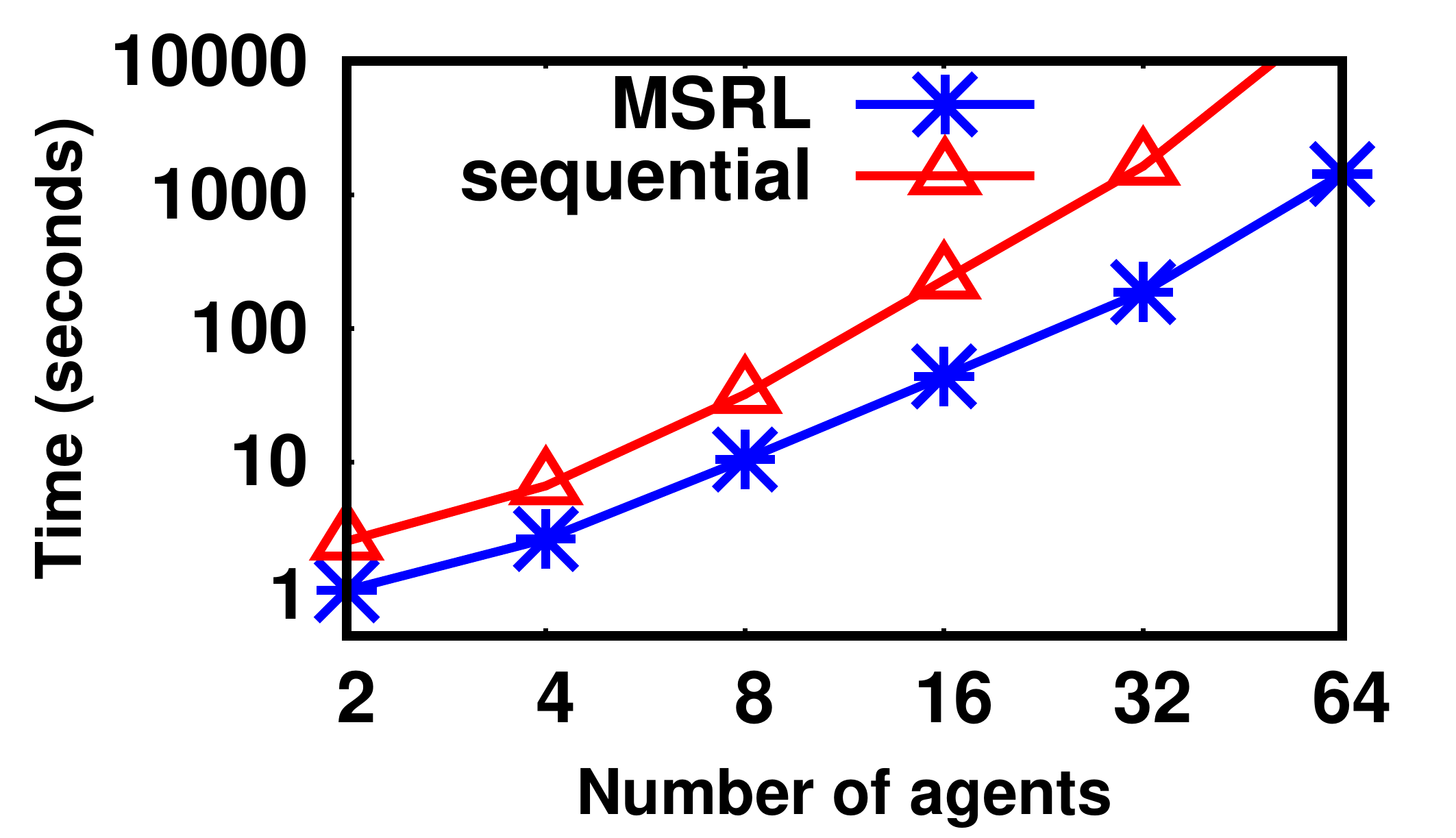}
    \caption{Training time per episode}\label{fig:ScaleMAPPO}
  \end{subfigure}
  \begin{subfigure}[t]{0.47\columnwidth}
    \includegraphics[width=\textwidth]{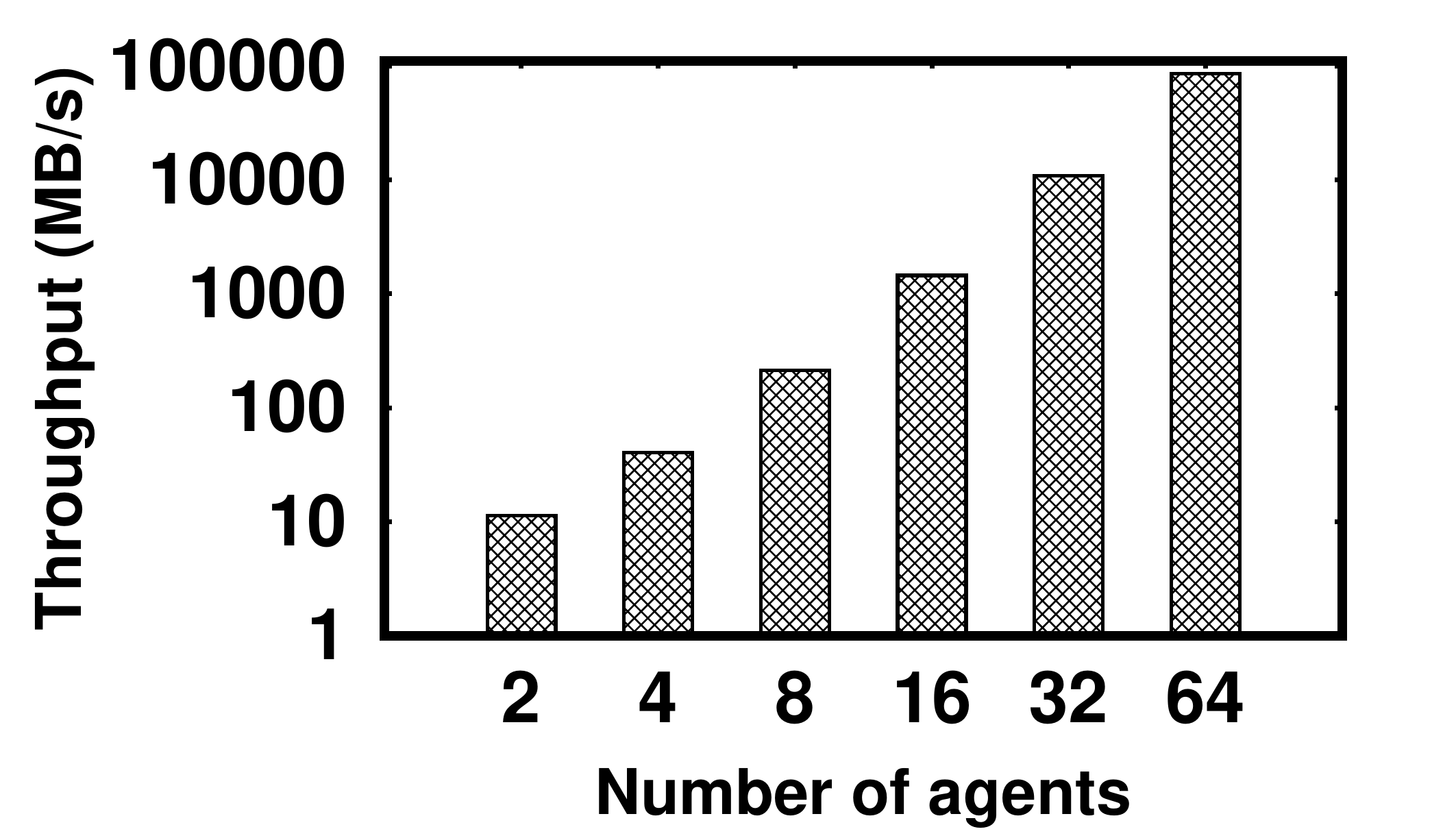}
    \caption{Training throughput}\label{fig:MAPPOThroughput}
  \end{subfigure}
  \caption{Scalability with agent count (MAPPO)}\label{fig:CloudMAPPO}
\end{figure}

\subsection{Scalability}
\label{sec:eval:scalability}

We explore how \sys{}'s design scales as we increase the number of deployed agents for a MARL algorithm and the number of environments (thus increasing the amount of training data). We want to validate if \sys{}'s approach introduces scalability bottlenecks.

\mypar{Agents} We increase the number of agents for the MAPPO MARL algorithm with the MPE \emph{simple spread}~\cite{DBLP:conf/nips/LoweWTHAM17} environment, where $n$~agents interact and learn to cover $n$~landmarks (rewards) while avoiding collisions (penalties). In addition to training local observations, agents must also train global observations on how far the closest agent is to each landmark. This results in O($n^3$)~observations with $n$~agents, thus quickly growing the computation complexity and GPU memory footprint~\cite{DBLP:conf/nips/LoweWTHAM17}. We deploy the MAPPO agents on the cloud cluster using \dpfive: each GPU on a worker trains 1~agent, and a dedicated worker executes all environments.

\F\ref{fig:ScaleMAPPO} shows the training time per episode for up to 64~GPUs against a sequential baseline (1~GPU). Due to its cubic complexity, the training time increases sharply both for the baseline and \sys. With distributed training, \sys's training time grows more slowly than the baseline: when training 32~agents, \sys improves the performance by 58$\times$; with 64~agents, the baseline exhausts GPU memory, while \sys trains one episode in 23.8\unit{mins}.

\F\ref{fig:MAPPOThroughput} compares the throughput when training different numbers of agents. The throughput is measured as the amount of data trained per second~(in MB/s). Increasing the agent number (\ie GPU devices) significantly improves throughput, and the margin becomes larger with more GPUs: the throughput when training 64~agents is over 7,600$\times$ higher than training 2~agents. This is because multiple GPUs train agents in parallel, and more agents result in more training data (\ie a larger size of observations) per GPU.

\begin{figure}[tb]
  \centering
  \includegraphics[width=0.9\columnwidth]{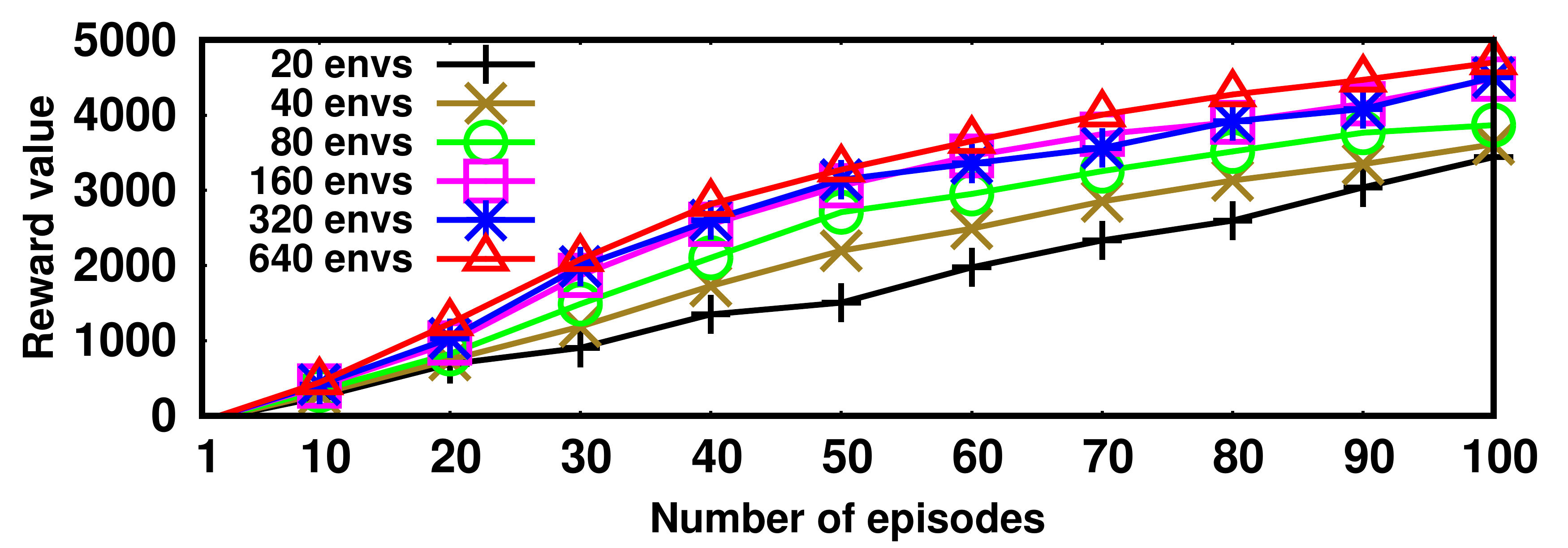}
  \caption{Statistical efficiency with environment count (PPO)}\label{fig:ScalePPOConvergence}
\end{figure}

\mypar{Environments} Next we increase the number of environments and observe the impact on the statistical efficiency, \ie the number of episodes to reach a given reward value. We assign 10~environments to each CPU and add more devices in the cloud cluster using \dptwo. We expect \sys to achieve higher statistical efficiency with more training data. 

In \F\ref{fig:ScalePPOConvergence}, we show the reward along with the number of episodes for different environment counts. As expected, increasing the number of environments leads to a higher reward value. As more CPUs execute environments, more trajectories are used per episode, achieving a higher reward.

\myparii{Conclusions:} \sys{}'s FDG abstraction does not deteriorate scalability. \sys scales with a large number of data-intensive agents, handling the increase in communication between fragments without bottlenecks. As expected, a larger amount of data generated by more environments also increases the statistical efficiency of the training process.



\section{Conclusions}
\label{sec:conclustion}

We described \emph{MindSpore Reinforcement Learning} ({\sys}), a system that supports the flexible parallelisation and distribution of RL algorithms on multiple devices. \sys separates the algorithm implementation from its execution through the abstraction of fragmented dataflow graphs~(FDGs), which enable \sys to support diverse distribution policies for allocating code fragments to devices. Our experimental results show the trade-offs when using different distribution policies.


\bibliographystyle{plain}

\end{document}